\documentclass[11pt]{article}

\usepackage[final]{acl}

\usepackage{times}
\usepackage{latexsym}
\usepackage[T1]{fontenc}
\usepackage[utf8]{inputenc}
\usepackage{microtype}
\usepackage{inconsolata}
\usepackage{graphicx}

\usepackage{stfloats}
\usepackage{multirow}
\usepackage{booktabs}
\usepackage{array}
\usepackage{makecell}
\usepackage{amsmath}
\usepackage{amssymb}

\title{Beyond Factual Accuracy: A Benchmark for Evaluating Perspective Alignment in LLMs}

\author{
  Sunghoon Kim \\
  Coxwave \\
  Seoul, Republic of Korea \\
  \texttt{sunghoon.kim@coxwave.com}
  \And
  Sangyeop Kim\thanks{Corresponding author.} \\
  Ewha Womans University \\
  Seoul, Republic of Korea \\
  \texttt{yeop@ewha.ac.kr}
}

\begin{document}
\maketitle

\begin{abstract}
Large Language Models often generate factually correct responses that fail to maintain user-requested perspectives, blending multiple perspectives and forcing users to filter through unwanted information—a perspective control failure that factual accuracy metrics cannot detect. We introduce \textbf{PerspectiveBench}, a benchmark quantifying perspective alignment by constructing scenarios where multiple valid reasoning paths coexist, and propose the \textbf{Perspective Alignment Score (PAS)} to measure adherence to specified perspectives. Experiments reveal that state-of-the-art models successfully include target content but fail to exclude non-requested perspectives. Providing documents through RAG improves factual grounding but does not resolve perspective drift, as models prioritize comprehensive coverage over strict user constraints. Code and data are available at \url{https://github.com/sunghoon014/PerspectiveBench}.

\end{abstract}

\section{Introduction}

People increasingly extract information directly through Large Language Models (LLMs) rather than navigating search results~\cite{llm-search}. LLMs process knowledge through reasoning—transforming data into insights and adapting it to context—enhancing the efficiency of information gathering.

However, this convenience introduces a critical challenge: overload from responses misaligned with user intent. Prior advances in information access have focused on \textbf{recall}—ensuring relevant content is retrieved and included—but as LLMs internalize web-scale knowledge, the bottleneck shifts to \textbf{precision}: selecting only what users actually requested. This shift exposes a structural bias in current alignment, where reward models systematically prefer longer, more comprehensive responses~\cite{deepseek-r1}, causing models to integrate all available knowledge rather than constraining outputs to the requested scope~\cite{mit2025cognitive}. This reveals a distinct alignment requirement: models must learn not only to produce valid content but also to withhold content beyond the user's requested scope—an ability orthogonal to factual correctness.

\begin{figure}[!htbp]
    \centering
    \includegraphics[width=0.9\linewidth]{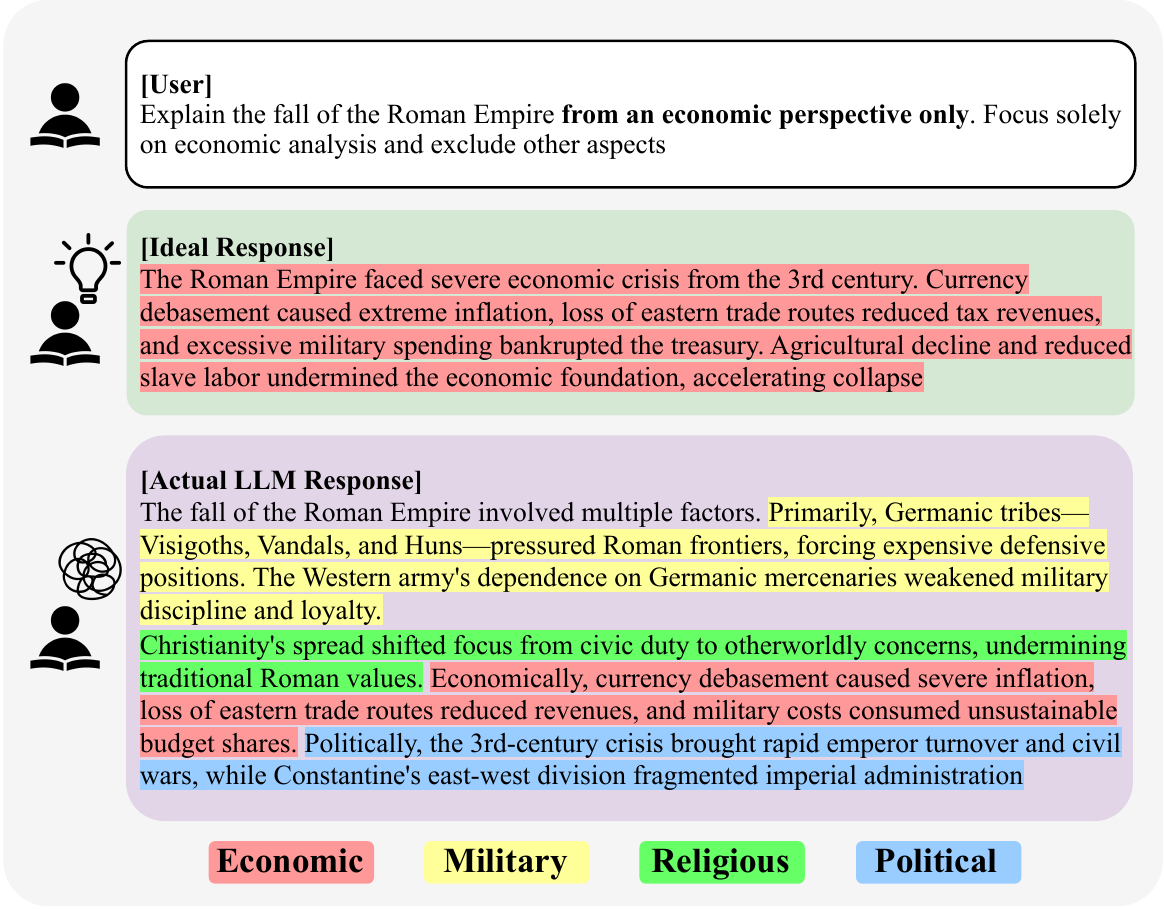}
    \caption{Perspective misalignment problem.}
    \label{fig:example}
\end{figure}

This overload manifests in two ways. First, LLMs fail to maintain the requested explanatory approach, or perspective, even when alternative accounts are also factually correct. Perspectives differ in the explanatory relation connecting the same premises and conclusion, rather than in subjective stance (Section~\ref{sec:Problem Formulation}). Consider a student asked to explain Rome's fall from an economic perspective only: the task requires a single causal account from economic factors alone (Figure~\ref{fig:example}). While an ideal response focuses exclusively on economic factors like currency debasement, actual responses typically lead with military or political explanations and then blend in economic and religious factors despite explicit constraints. Second, providing perspective-specific documents through RAG fails to resolve this; models still blend in alternative perspectives from training knowledge rather than adhering to provided materials~\cite{conflict}. Both problems reveal the same core limitation: models excel at determining what is factually correct but struggle to calibrate what users actually want within that factual space.

Current benchmarks rarely address this gap. Hallucination detection targets false information, while RAG evaluation verifies output quality against retrieved content~\cite{benchmark-rageval}—both concentrate on factual correctness without assessing perspective adherence. Instruction-following and personalization benchmarks evaluate constraint compliance or steering toward a target, while perspective-recognition benchmarks identify viewpoints. These evaluations do not directly measure exclusion of factually valid reasoning outside the requested perspective (Section~\ref{sec:related}). A response may therefore be factually accurate and topically relevant while still failing to maintain the requested perspective.

We propose \textbf{PerspectiveBench}, which evaluates how precisely LLMs control response perspectives. The benchmark constructs scenarios where multiple valid reasoning paths coexist for a single question, then measures whether models stay within the specific perspective users request. We introduce the \textbf{Perspective Alignment Score (PAS)} to quantify this alignment, while using the established Faithfulness for document adherence in RAG settings.

Our main contributions are as follows:
\begingroup
\makeatletter
\def\@listi{\leftmargin\leftmargini
    \topsep 2pt \partopsep 0pt \itemsep 1pt \parsep 0pt}
\makeatother
\begin{itemize}
    \item We identify \textbf{perspective misalignment} as a failure mode and frame it as a shift from recall- to precision-oriented evaluation.
    \item We introduce \textbf{PerspectiveBench} and the \textbf{Perspective Alignment Score (PAS)}—the first benchmark and metric that jointly measure selective exclusion among semantically relevant perspectives, quantifying whether models cover target reasoning paths while suppressing factually valid but off-scope content.

    \item We show that state-of-the-art LLMs include target content but fail to exclude off-scope content, and that ideal RAG context improves grounding but not perspective control—revealing inclusion and exclusion as decoupled capabilities.

\end{itemize}
\endgroup

\section{PerspectiveBench}
Our pipeline combines automated knowledge-graph generation with rigorous human validation of the resulting QA pairs, ensuring quality with minimal annotation cost; inter-rater reliability is reported in Appendix~\ref{sec:validity}, and all prompts used in pipeline construction are provided in Appendix~\ref{sec:app-prompts-templates}.

\begin{figure*}[!th]
  \centering
  \includegraphics[width=\linewidth]{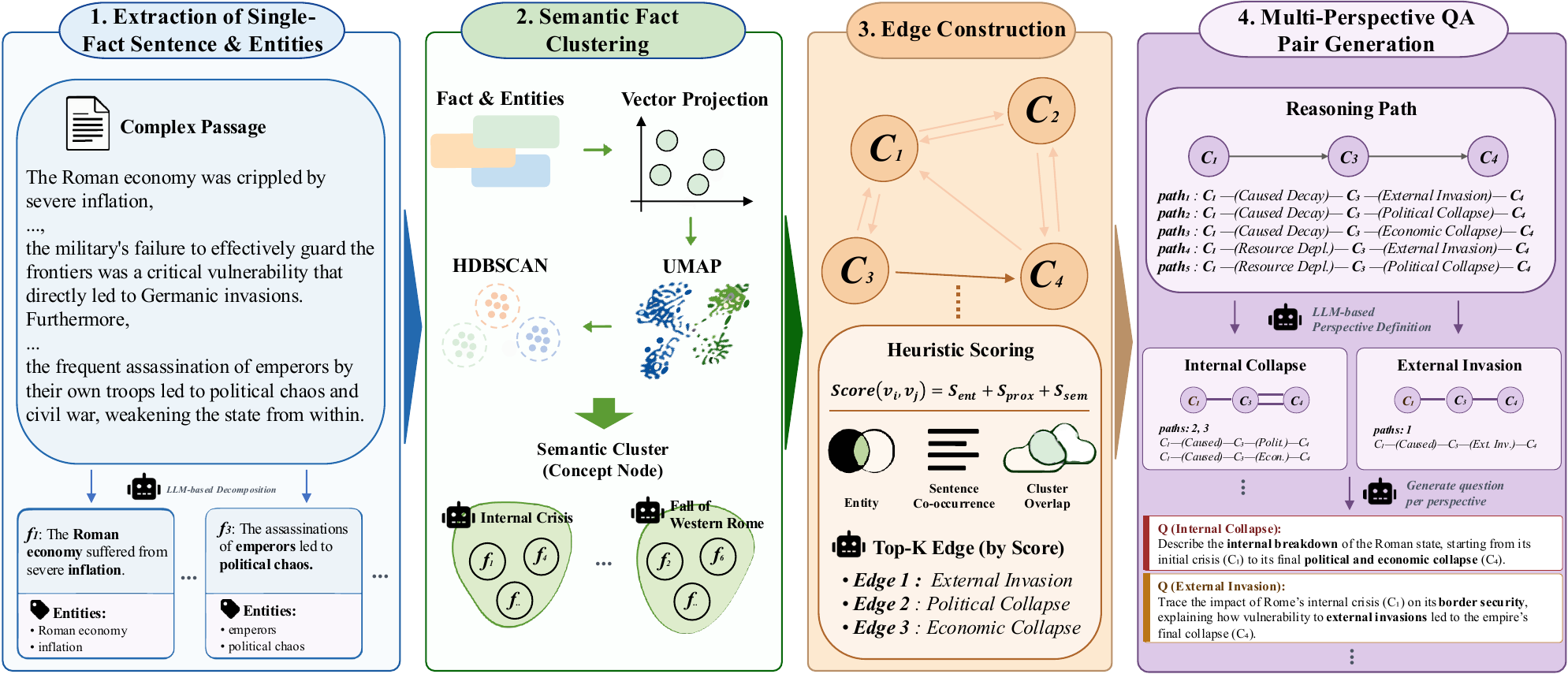}
  \caption{The PerspectiveBench data construction pipeline: From atomic fact extraction and semantic clustering to graph-based reasoning path and perspective-targeted QA generation.}
  \label{fig:main}
\end{figure*}

\subsection{Problem Formulation}
\label{sec:Problem Formulation}

For a query $q$, multiple valid reasoning paths may exist: $\Pi(q) = \{\pi_1, \pi_2, \ldots, \pi_n\}$. Each path $\pi_i$ represents a coherent explanatory chain from premises to conclusion.
Given a user-specified perspective $\mathcal{P}$—an explanatory approach defined by specific reasoning strategies—these paths partition into:
$$
\Pi(q) = \Pi_{\mathcal{P}}^{\text{target}}(q) \cup \Pi_{\mathcal{P}}^{\text{distractor}}(q)
$$
where $\Pi_{\mathcal{P}}^{\text{target}}(q)$ contains paths aligned with perspective $\mathcal{P}$, and $\Pi_{\mathcal{P}}^{\text{distractor}}(q)$ contains factually valid paths misaligned with $\mathcal{P}$.

Perspective-controlled models must generate responses that maximize alignment with $\mathcal{P}$:
$$
\hat{a} = \operatorname*{argmax}_{a:\, \text{supp}(a) \subseteq \Pi_{\mathcal{P}}^{\text{target}}(q)} P(a \mid q, \mathcal{P})
$$
where $\text{supp}(a)$ denotes the reasoning paths supporting response $a$. This requires including covering target paths while excluding distractors despite their availability. The challenge lies in distinguishing $\Pi_{\mathcal{P}}^{\text{distractor}}(q)$—paths that are factually correct and semantically relevant, yet employ misaligned explanatory approaches. Such paths constitute hard negatives from a perspective control standpoint, as they cannot be filtered by factual accuracy metrics alone~\cite{ragas}.

\subsection{Source Dataset}
\label{sec:Source Dataset}

The benchmark requires data where multiple valid perspectives coexist for single topics. We construct a controlled testbed from OpenStax textbooks~\cite{openstax}, which provide peer-reviewed academic content while preserving perspective diversity. This control helps distinguish exclusion failure from factual error or retrieval error. We select three domains—Physics, Biology, and History—where single questions naturally admit multiple valid perspectives (e.g., mechanical vs.\ thermodynamic frames for energy conservation; economic, military, or political accounts of Rome's fall); per-domain rationales are in Appendix~\ref{sec:app-domains}. Preprocessing splits texts at paragraph breaks with a 1,024-token cap, yielding 706 chunks.

\subsection{Benchmark Construction}
\label{section:construction}

\subsubsection{Extraction of Single-Fact Sentence \& Entities}
\label{sec:Extraction of Single Fact Sentences}

Complex textual passages often combine multiple facts within single sentences, making it difficult to distinguish specific perspectives. The first construction stage decomposes text chunks $T$ from Section~\ref{sec:Source Dataset} into single-fact sentences $\mathcal{F} = \{f_1, f_2, \ldots, f_n\}$, where each $f_i$ contains exactly one discrete piece of information.

For example, the passage in Figure~\ref{fig:main}—``The Roman economy was crippled by severe inflation, and the military's failure to guard frontiers directly led to Germanic invasions''—is decomposed into $f_1$ = ``The Roman economy suffered from severe inflation'' and $f_2$ = ``The military's failure to guard frontiers led to Germanic invasions.'' Each fact is then assigned to a specific perspective during clustering (Section~\ref{sec:Semantic Fact Clustering}).

LLMs perform this extraction automatically while simultaneously identifying key entities $e_i$ for each fact $f_i$.
For instance, $f_1$ yields $e_1 = \{$``Roman economy'', ``inflation''$\}$, and $f_2$ yields $e_2 = \{$``military'', ``frontiers'', ``Germanic invasions''$\}$.
These entities serve dual purposes: they constitute the semantic core of each fact and provide signals for computing relevance scores between node pairs during edge construction. Additionally, these single-fact sentences later serve as ground-truth documents during evaluation, ensuring factually complete and perspective-aligned context.

\subsubsection{Semantic Fact Clustering}
\label{sec:Semantic Fact Clustering}

The single-fact sentences in $\mathcal{F}$ are too fine-grained to represent high-level perspectives. We group semantically related facts into cohesive clusters $\{C_1, C_2, \ldots, C_k\}$ that form semantic nodes $\mathcal{V}$ in our knowledge graph.

We apply a three-stage pipeline for density-based clustering. First, F2LLM-4B~\cite{F2LLM}, validated on MTEB clustering benchmarks~\cite{mteb}, projects each fact $f_i$ into a high-dimensional vector space. Second, UMAP~\cite{umap} compresses these embeddings into lower dimensions while preserving local semantic structure, improving clustering efficiency. Finally, HDBSCAN~\cite{hdbscan} identifies dense regions to form clusters without requiring predetermined cluster counts. Critically, HDBSCAN automatically excludes low-density outliers as noise, filtering out ambiguous or irrelevant sentences at the construction stage and ensuring semantic purity of resulting nodes.

Each cluster $C_j$ requires an explicit concept label to function as a graph node. LLMs generate representative summaries $s_j$ that synthesize semantic relationships within each cluster. For instance, cluster $C_j = \{$\text``The Roman economy suffered from severe inflation''\}, \text{``Constant civil wars weakened the state''}$\}$ is summarized as $s_j = $ ``Internal Crisis,'' which synthesizes economic collapse and political instability as internal vulnerabilities. Each summary $s_j$, paired with its constituent facts $C_j$, becomes a semantic node $v_j \in \mathcal{V}$ for subsequent edge construction.

\subsubsection{Edge Construction}
\label{sec:Edge}

This stage constructs directed explanatory edges between semantic nodes to complete the knowledge graph $G=(\mathcal{V}, \mathcal{E})$. Since evaluating all node pairs through LLMs incurs $O(N^2)$ cost, we introduce a weighted heuristic $\text{Score}(v_i, v_j) = S_{\text{ent}} + S_{\text{prox}} + S_{\text{sem}}$, where $S_{\text{ent}}$ rewards entity overlap between clusters, $S_{\text{prox}}$ rewards sentence-level co-occurrence, and $S_{\text{sem}}$ rewards bidirectional soft-cluster overlap from HDBSCAN. Full point assignments are in Appendix~\ref{sec:app-edge-scoring}.

For each top-scoring pair, an LLM identifies all explanatory relationships between the clusters and generates one labeled edge per relationship. A pair may carry multiple edges with distinct labels. In Figure~\ref{fig:main}, from ``Military Decay'' (Cluster B) to ``Fall of Western Rome'' (Cluster C), the LLM produces three edges labeled ``External Invasion,'' ``Political Collapse,'' and ``Economic Collapse,'' which serve as the basis for distinct perspectives.

\subsubsection{Multi-Perspective Question-Answer Pair Generation}
\label{sec:Perspective}

Using the knowledge graph $G=(\mathcal{V}, \mathcal{E})$, we generate QA pairs in three stages. A reasoning path $\pi = (v_0 \rightarrow v_1 \rightarrow \cdots \rightarrow v_k)$ traverses directed edges $(v_i, v_{i+1}) \in \mathcal{E}$ with their relationship labels, ranging from 1-hop ($v_s \rightarrow v_e$) to 3-hop. For each endpoint pair $(v_s, v_e)$, we group the valid paths $\Pi(v_s \rightarrow v_e)$ into perspectives $\mathcal{P}_i$, where each perspective groups paths that share the same explanatory focus (e.g., causal, comparative, structural). In Figure~\ref{fig:main}, paths through ``Political Collapse'' and ``Economic Collapse'' share an internal explanatory focus, forming perspective $\mathcal{P}_{\text{internal}}$, while ``External Invasion'' constitutes a separate perspective. For each perspective, an LLM generates a question with conceptual hints toward its target paths without revealing edge labels; the question for $\mathcal{P}_{\text{internal}}$ asks about ``\textit{internal breakdown}'' focusing on ``\textit{political and economic collapse}.'' Ground-truth documents per question are single-fact sentences from clusters along target paths $\Pi_{\mathcal{P}}^{\text{target}}$, as context for ideal-retrieval evaluation. Step-by-step formalism is in Appendix~\ref{sec:app-qa-gen}.

\subsection{Dataset Characteristics}

\paragraph{Statistics.}
Table~\ref{tab:benchmark-stats} shows 1,024 questions across three domains. Each question corresponds to paths $\Pi(v_s \rightarrow v_e)$ with an average of 5.24 total paths, partitioned into 2.16 target paths $\Pi_{\mathcal{P}}^{\text{target}}$ and 3.08 distractor paths $\Pi_{\mathcal{P}}^{\text{distractor}}$—the latter being factually valid but employing misaligned explanatory approaches. Models must selectively include targets while excluding distractors, the core challenge of perspective control. Per-domain perspective-type distributions are reported in Appendix~\ref{sec:app-theme}.

\begin{table}[htbp]
    \caption{Benchmark statistics by domain and hop.}
    \label{tab:benchmark-stats}
    \centering
    \footnotesize
    \renewcommand{\arraystretch}{0.85}
    \setlength{\tabcolsep}{4pt}
    \resizebox{\columnwidth}{!}{%
    \begin{tabular}{llrrrr}
        \toprule
        \textbf{Domain} & \textbf{Hop} & \textbf{Avg. Total} & \textbf{Avg. Target} & \textbf{Avg. Distractor} & \textbf{Num. Qs} \\
        \midrule
        \multirow{3}{*}{\textbf{Physics}} & 1 & 2.77 & 1.06 & 1.71 & 114 \\
                                          & 2 & 6.20 & 2.60 & 3.60 & 161 \\
                                          & 3 & 7.47 & 3.60 & 3.86 & 58 \\
        \midrule
        \multirow{3}{*}{\textbf{Biology}} & 1 & 3.34 & 1.23 & 2.11 & 103 \\
                                          & 2 & 5.50 & 2.35 & 3.15 & 164 \\
                                          & 3 & 7.25 & 2.92 & 4.33 & 61 \\
        \midrule
        \multirow{3}{*}{\textbf{History}} & 1 & 4.25 & 1.30 & 2.95 & 146 \\
                                          & 2 & 5.71 & 2.47 & 3.24 & 161 \\
                                          & 3 & 7.00 & 3.21 & 3.79 & 56 \\
        \midrule
        \textbf{Avg. / Total} & & \textbf{5.24} & \textbf{2.16} & \textbf{3.08} & \textbf{1,024} \\
        \bottomrule
    \end{tabular}%
    }
\end{table}

\begingroup
\setlength{\intextsep}{6pt plus 1pt minus 1pt}
\begin{figure}[htbp]
    \centering
    \captionsetup{skip=4pt}
    \includegraphics[width=\linewidth]{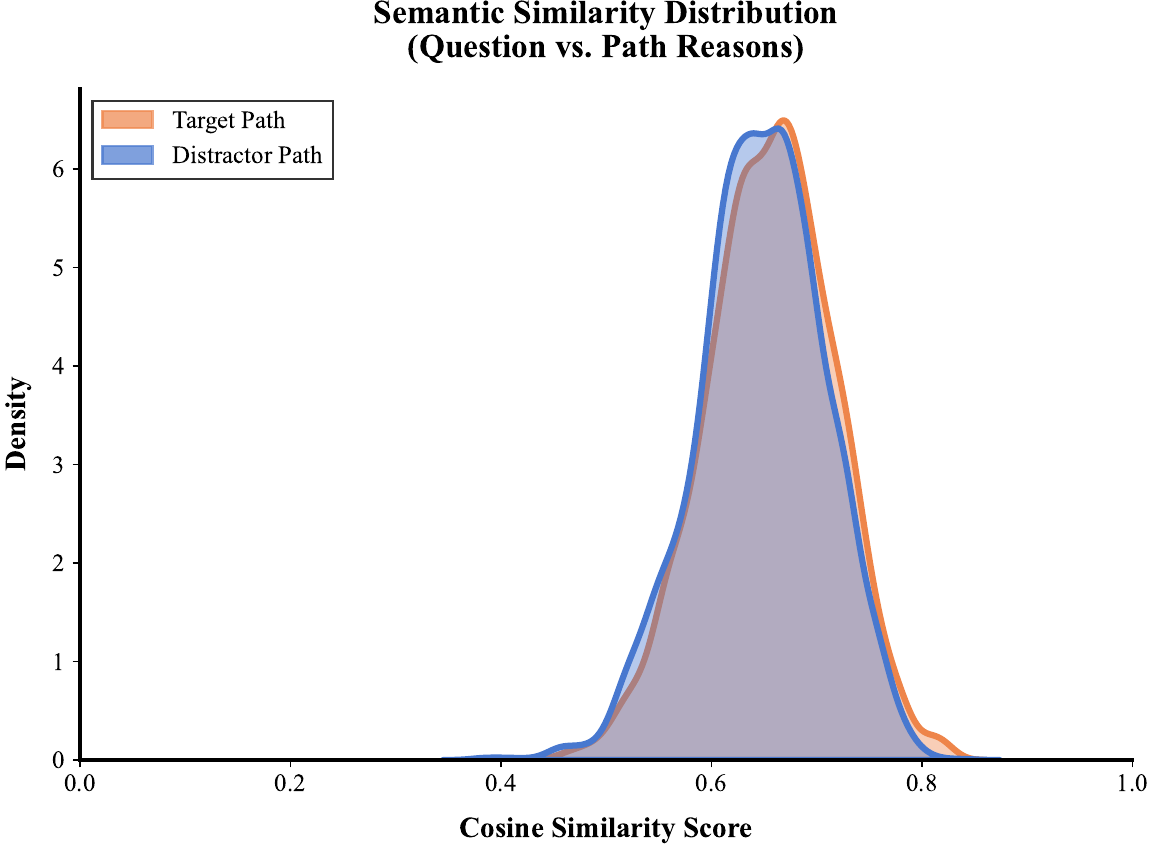}
    \caption{\textbf{Semantic similarity (bge-m3).} Target and distractor paths overlap.}
    \label{fig:sim_dist}
\end{figure}
\endgroup

\paragraph{Semantic Ambiguity of Distractors.}
\label{sec:ambiguity}
Figure~\ref{fig:sim_dist} shows that target and distractor paths exhibit nearly identical semantic similarity to queries, indicating that distractors are not off-topic noise but semantically relevant alternatives differing only in perspective. This overlap is a designed property of the benchmark: since distractor paths are derived from graph edges with distinct explanatory relationships (Section~\ref{sec:Edge}), their topical proximity is precisely what renders perspective control non-trivial. Vector-space proximity cannot separate them, making them hard negatives that simple relevance scoring cannot filter—benchmarks without this property reduce to topic classification.

\begin{table*}[!t]
    \caption{Results by model and hop. Overall scores are weighted by questions per hop; \#Tok denotes mean generated tokens. Bold and underline denote the best and second-best scores per column.}
    \label{tab:main-pfs-cfs-results}
    \centering
    \small
    \renewcommand{\arraystretch}{1.25}
    \setlength{\tabcolsep}{4.2pt}
    \resizebox{\textwidth}{!}{
    \begin{tabular}{ll|rrrrr|rrrrr|rrrrr|rrrrr}
        \toprule
        \textbf{Model} & \textbf{Setting}
            & \multicolumn{5}{c|}{\textbf{1-hop}}
            & \multicolumn{5}{c|}{\textbf{2-hop}}
            & \multicolumn{5}{c|}{\textbf{3-hop}}
            & \multicolumn{5}{c}{\textbf{Overall}} \\
        \cmidrule(lr){3-7} \cmidrule(lr){8-12} \cmidrule(lr){13-17} \cmidrule(lr){18-22}
        &
            & \textbf{$P_{\text{Prec}}$} & \textbf{$P_{\text{Rec}}$} & \textbf{PAS} & \textbf{Faithfulness} & \textbf{\#Tok}
            & \textbf{$P_{\text{Prec}}$} & \textbf{$P_{\text{Rec}}$} & \textbf{PAS} & \textbf{Faithfulness} & \textbf{\#Tok}
            & \textbf{$P_{\text{Prec}}$} & \textbf{$P_{\text{Rec}}$} & \textbf{PAS} & \textbf{Faithfulness} & \textbf{\#Tok}
            & \textbf{$P_{\text{Prec}}$} & \textbf{$P_{\text{Rec}}$} & \textbf{PAS} & \textbf{Faithfulness} & \textbf{\#Tok} \\
        \specialrule{1pt}{1pt}{1pt}

        \multirow{2}{*}{GPT-5} & Base
            & 44.37 & 92.50 & 56.13 & 33.06 & 464.92
            & 37.33 & 58.66 & 42.13 & 35.56 & 489.89
            & 35.90 & 46.79 & 37.27 & 37.26 & 524.71
            & 39.58 & 68.63 & 46.26 & 34.96 & 486.99 \\
        & Oracle
            & 48.75 & \underline{97.50} & 61.55 & \textbf{68.48} & 206.17
            & \textbf{49.00} & \textbf{71.33} & \textbf{54.41} & \underline{65.84} & 241.76
            & 48.07 & 60.89 & 49.74 & \underline{70.23} & 288.52
            & \textbf{48.75} & \textbf{78.82} & \textbf{56.14} & \textbf{67.53} & 237.13 \\
        \midrule
        \multirow{2}{*}{GPT-4.1} & Base
            & \underline{51.50} & 96.67 & \underline{63.80} & 51.88 & 301.58
            & 39.33 & 44.00 & 38.29 & 48.16 & 331.95
            & 31.41 & 39.74 & 32.75 & 51.92 & 361.98
            & 42.29 & 61.94 & 46.39 & 50.12 & 326.32 \\
        & Oracle
            & 50.16 & \underline{97.50} & 62.68 & 66.64 & 270.59
            & 39.00 & 56.00 & 43.24 & 62.24 & 286.85
            & 46.31 & \textbf{65.06} & 48.93 & 68.20 & 351.52
            & 44.21 & 72.26 & 51.10 & 64.82 & 292.14 \\
        \specialrule{1pt}{1pt}{1pt}
        \multirow{2}{*}{Claude-Sonnet-4.5} & Base
            & 43.87 & 90.83 & 56.12 & 44.67 & 405.14
            & 35.00 & 49.33 & 37.06 & 42.10 & 413.23
            & 34.62 & 44.87 & 37.19 & 44.71 & 433.09
            & 38.08 & 63.28 & 43.84 & 43.46 & 413.76 \\
        & Oracle
            & 44.79 & 95.00 & 57.62 & 65.40 & 365.86
            & \underline{48.00} & \underline{69.33} & 52.55 & 62.10 & 373.01
            & \textbf{51.92} & 58.97 & \textbf{51.47} & 63.84 & 405.85
            & \underline{47.53} & 76.66 & 54.16 & 63.57 & 376.09 \\
        \midrule
        \multirow{2}{*}{Claude-Sonnet-4} & Base
            & 43.25 & 92.50 & 56.71 & 37.07 & 390.88
            & 34.33 & 52.66 & 37.75 & 36.44 & 415.36
            & 29.80 & 43.59 & 33.28 & 44.25 & 406.52
            & 36.72 & 65.23 & 43.71 & 38.00 & 405.17 \\
        & Oracle
            & 45.45 & 94.17 & 57.88 & 56.73 & 320.28
            & 33.00 & 60.00 & 40.80 & 59.75 & 330.25
            & 32.69 & 50.00 & 37.90 & 63.40 & 363.12
            & 37.36 & 70.40 & 46.36 & 59.30 & 332.33 \\
        \specialrule{1pt}{1pt}{1pt}
        \multirow{2}{*}{Qwen3-32B} & Base
            & 46.67 & 92.50 & 58.29 & 37.55 & 275.16
            & 40.00 & 62.00 & 45.26 & 41.92 & 273.33
            & 25.64 & 33.97 & 27.75 & 41.74 & 283.70
            & 39.91 & 68.02 & 46.89 & 40.34 & 275.75 \\
        & Oracle
            & 46.87 & 94.16 & 58.29 & 49.13 & 227.52
            & 46.66 & \textbf{71.33} & \underline{53.10} & 53.10 & 240.89
            & 48.07 & \underline{62.50} & \underline{50.98} & 51.09 & 242.61
            & 46.98 & \underline{77.91} & \underline{54.58} & 51.35 & 236.44 \\
        \midrule
        \multirow{2}{*}{Qwen3-14B} & Base
            & \textbf{52.50} &\textbf{ 98.33} &\textbf{ 64.55} & 31.60 & 170.73
            & 29.33 & 42.66 & 32.95 & 42.42 & 218.68
            & 32.05 & 41.66 & 32.98 & 54.68 & 214.30
            & 38.01 & 62.22 & 44.16 & 40.68 & 200.93 \\
        & Oracle
            & 47.17 & 90.83 & 57.60 & 47.92 & 173.45
            & 39.33 & 57.33 & 44.32 & 60.63 & 199.17
            & 36.54 & 47.44 & 38.16 & 62.80 & 231.44
            & 41.63 & 67.52 & 47.97 & 56.50 & 195.57 \\
        \midrule
        \multirow{2}{*}{Qwen3-8B} & Base
            & 48.54 & 96.66 & 60.95 & 28.06 & 205.55
            & 42.67 & 56.67 & 44.57 & 41.68 & 217.62
            & 24.67 & 27.89 & 24.00 & 60.73 & 218.11
            & 41.67 & 65.93 & 46.86 & 40.11 & 213.43 \\
        & Oracle
            & 49.50 & 95.00 & 61.55 & 43.39 & 180.45
            & 40.33 & 68.00 & 48.02 & 60.57 & 216.72
            & 28.84 & 41.67 & 30.81 & 66.04 & 226.85
            & 41.62 & 73.07 & 49.88 & 55.41 & 205.59 \\
        \specialrule{1pt}{1pt}{1pt}
        \multirow{2}{*}{Gemma3-27B} & Base
            & 43.54 & 93.33 & 56.04 & 50.31 & 431.02
            & 36.66 & 53.33 & 40.26 & 45.91 & 470.61
            & 33.33 & 39.42 & 32.21 & 53.13 & 495.17
            & 38.53 & 65.13 & 44.48 & 48.70 & 460.77 \\
        & Oracle
            & 45.54 & 91.25 & 57.13 & \underline{67.40} & 265.64
            & 47.00 & 66.67 & 51.22 & 63.95 & 279.98
            & 45.51 & 56.41 & 46.67 & 68.20 & 315.87
            & 46.23 & 73.63 & 52.54 & 65.90 & 281.03 \\
        \midrule
        \multirow{2}{*}{Gemma3-12B} & Base
            & 46.04 & 90.83 & 57.30 & 25.62 & 517.60
            & 39.33 & 50.00 & 39.07 & 41.50 & 547.40
            & 32.69 & 39.10 & 32.46 & 55.41 & 578.44
            & 40.57 & 62.61 & 44.40 & 38.25 & 542.14 \\
        & Oracle
            & 44.75 & 91.67 & 57.10 & 61.59 & 272.97
            & 47.00 & 52.67 & 46.80 & \textbf{66.26} & 291.37
            & 45.51 & 43.91 & 41.39 & \textbf{77.66} & 385.85
            & 45.95 & 65.00 & 49.53 & \underline{66.55} & 300.99 \\
        \midrule
        \multirow{2}{*}{Gemma3-4B} & Base
            & 30.63 & 61.67 & 38.33 & 25.97 & 497.97
            & 27.00 & 33.33 & 26.27 & 40.04 & 473.27
            & 19.87 & 25.64 & 20.98 & 55.96 & 514.30
            & 27.07 & 42.06 & 29.64 & 37.77 & 489.04 \\
        & Oracle
            & 36.04 & 60.83 & 42.33 & 46.77 & 280.47
            & 30.00 & 48.00 & 33.62 & 56.96 & 300.07
            & 31.41 & 44.87 & 34.62 & 61.87 & 318.96
            & 32.38 & 52.01 & 36.88 & 54.19 & 296.35 \\
        \specialrule{1pt}{1pt}{1pt}
    \end{tabular}
    }
\end{table*}

\section{Experiment}
\subsection{Experimental Setup}

We evaluate frontier LLMs from GPT, Claude, Qwen3, and Gemma3 families on PerspectiveBench under two conditions. The \textbf{Base} setting provides only questions, requiring models to maintain requested perspectives using parametric knowledge alone. The \textbf{Oracle} setting simulates ideal RAG by providing ground-truth documents supporting the requested perspective, testing whether optimal context helps models maintain focus or whether they still blend in alternative perspectives.
We use Gemini-2.5-Pro as the LLM-as-a-Judge. Full setup details are in Appendix~\ref{sec:app-exp-setup}.

\subsection{Evaluation Metrics}
\label{sec:eval-metrics}
Following CheckEval~\cite{checkeval}, we adopt an LLM-as-a-Judge evaluation framework that formulates the task as checklist-based verification. We compute two metrics from these judgments.

\textbf{Perspective Alignment Score (PAS)} measures whether a response covers target reasoning paths while excluding distractor paths—a graph-structured generalization of precision/recall. For each question, target paths $T \subseteq \Pi(q)$ denote the complete set of paths matching the requested perspective. Instead of a generic scoring rubric, we use these ground-truth paths from our knowledge graph as a dynamic checklist. The judge iterates through each path $\pi \in \Pi(q)$ and determines whether the response semantically reproduces all of its constituent edges, yielding the set of identified paths:
$$
A = \{\pi \in \Pi(q) : \text{Judge}(\pi, \text{response}) = \text{present}\}.
$$
Perspective precision $P_{\text{Prec}} = |A \cap T| / |A|$ is the proportion of included paths that are targets; Perspective recall $P_{\text{Rec}} = |A \cap T| / |T|$ is target path coverage. PAS is their harmonic mean.

\textbf{Faithfulness} adopts the standard claim-level verification metric~\cite{ragas}: the judge decomposes each response into discrete claims and verifies each against the ground-truth single-fact sentences from the target paths. Faithfulness is the proportion of supported claims. In Base settings, where no context is provided, these sentences are used only post-hoc, so the metric reflects factual coverage rather than grounding in the provided context.

To validate the PAS judge, human annotators evaluated responses using the same path checklist as the judge. Agreement over 220 path-level Yes/No decisions was 82.7\% (Cohen's $\kappa=0.64$), and the Spearman correlation between human and judge PAS scores at the response level was $\rho=0.71$. We re-scored PAS on the same responses from three models with a judge from a different model family (GPT-5.6 Terra), using the same prompt and output format; path-level agreement with the Gemini judge was 82.3\% ($\kappa=0.645$), similar to human--judge agreement (Appendix~\ref{sec:app-judge-robustness}).

\subsection{Results}
\label{sec:results}
\subsubsection{Models Include but Fail to Exclude}

Table~\ref{tab:main-pfs-cfs-results} reveals a fundamental failure in maintaining focus across all state-of-the-art models. In the Base setting, overall PAS ranges from 29.6 to 46.9. On 1-hop questions, most models achieve high perspective recall but critically low precision. Many models remain close to or below the cover-all precision of 41.4, and none reaches its PAS of 57.6 (Appendix~\ref{sec:app-cover-all}). In Base, a missed target path may reflect a knowledge gap or a failure to stay in scope; Oracle provides the target facts. Partial credit substantially reduces the 1-hop to 3-hop recall drop for Oracle responses from three models, while precision remains low (Section~\ref{sec:scoring-artifacts}). High recall on 1-hop questions shows that most models can include relevant information, while low precision across hops indicates difficulty in excluding non-requested perspectives. They default to comprehensive ``knowledge dumps'' rather than focused answers, leaving the filtering to the user (see Appendix~\ref{sec:app-cost} for an extended discussion).

\subsubsection{Oracle Improves Grounding but Not Selectivity}

As shown in Table~\ref{tab:main-pfs-cfs-results}, the Oracle setting, simulating ideal RAG with ground-truth documents, substantially improves Faithfulness by 46.5\% on average, indicating models successfully utilize provided context. With ground truth expanded to all graph sentences, including distractor facts, the Base--Oracle Faithfulness gap is largely preserved for four models (Appendix~\ref{sec:app-expanded-gt}). However, perspective control shows limited improvement. For 1-hop reasoning, Oracle provides minimal gains, with most models showing no meaningful precision improvement. At higher hop levels, Oracle shows relatively larger improvements, suggesting RAG helps more when parametric knowledge becomes uncertain. Yet absolute performance remains low, indicating RAG assistance alone cannot solve the focus problem.

\subsubsection{Conciseness Does Not Imply Selectivity}

Comparing Base and Oracle settings in Table~\ref{tab:main-pfs-cfs-results}, Oracle reduces average tokens by 29.5\% while substantially improving Faithfulness. Models become more concise and factually grounded under Oracle context. However, this conciseness does not yield selectivity: PAS shows minimal improvement, indicating that models compress multiple perspectives into shorter outputs rather than excluding alternatives. Conciseness and selective exclusion are orthogonal capabilities—a shorter response can still contain factually valid content from non-target perspectives. Since Oracle represents ideal retrieval, realistic RAG systems with imperfect retrieval will likely face even greater alignment challenges.

\section{Analysis}

\subsection{Prompting Fails at Selective Exclusion}
\label{sec:prompting}

To distinguish whether perspective control failure stems from instruction-following deficiency or inherent limitation, we compare three prompting strategies: \textbf{Standard Zero-shot}, \textbf{One-Shot Demonstration}, and \textbf{Negative Constraints}; full prompt templates are in Appendix~\ref{sec:app-prompts}.

\begin{table}[htbp]
    \caption{Prompting results. One-Shot and Negative Constraints: mean over questions. Best scores per model are bold; second-best are underlined.}
    \label{tab:ablation-study}
    \centering
    \small
    \renewcommand{\arraystretch}{0.9}
    \resizebox{\columnwidth}{!}{
    \begin{tabular}{lrrrrr}
        \toprule
        \textbf{Model / Strategy} & \textbf{$P_{\text{Prec}}$} & \textbf{$P_{\text{Rec}}$} & \textbf{PAS} & \textbf{Faithfulness} & \textbf{\#Tok} \\
        \specialrule{1pt}{1pt}{1pt}
        \multicolumn{6}{l}{\textit{\textbf{GPT-5}}} \\
        \hspace{2mm} Oracle & 48.75 & \underline{78.82} & \underline{56.14} & \textbf{67.53} & 237.13 \\
        \hspace{2mm} + One-Shot & \underline{49.18} & 78.02 & 55.95 & 63.46 & 249.55 \\
        \hspace{2mm} + Negative Constraints & \textbf{52.29} & \textbf{80.40} & \textbf{58.16} & \underline{66.43} & 254.86 \\
        \midrule
        \multicolumn{6}{l}{\textit{\textbf{Claude-Sonnet-4.5}}} \\
        \hspace{2mm} Oracle & \textbf{47.53} & \underline{76.66} & \textbf{54.16} & \underline{63.57} & 376.09 \\
        \hspace{2mm} + One-Shot & 43.05 & \textbf{77.75} & \underline{51.81} & \textbf{65.06} & 346.24 \\
        \hspace{2mm} + Negative Constraints & \underline{44.14} & 73.81 & 51.48 & 62.45 & 399.01 \\
        \midrule
        \multicolumn{6}{l}{\textit{\textbf{Qwen3-32B}}} \\
        \hspace{2mm} Oracle & \textbf{46.98} & \textbf{77.91} & \textbf{54.58} & \textbf{51.35} & 236.44 \\
        \hspace{2mm} + One-Shot & 40.66 & 68.86 & 47.34 & 48.47 & 240.55 \\
        \hspace{2mm} + Negative Constraints & \underline{44.96} & \underline{75.64} & \underline{52.17} & \underline{48.93} & 240.07 \\
        \midrule
        \multicolumn{6}{l}{\textit{\textbf{Gemma3-27B}}} \\
        \hspace{2mm} Oracle & \textbf{46.23} & \textbf{73.63} & \textbf{52.54} & \textbf{65.90} & 281.03 \\
        \hspace{2mm} + One-Shot & 42.80 & 67.77 & 48.71 & 60.76 & 266.55 \\
        \hspace{2mm} + Negative Constraints & \underline{45.40} & \underline{68.86} & \underline{50.72} & \underline{62.61} & 291.59 \\
        \bottomrule
    \end{tabular}
    }
\end{table}

Table~\ref{tab:ablation-study} shows that enhanced prompting fails to resolve perspective control. One-Shot demonstrations are ineffective: GPT-5 shows negligible improvement (+0.43\%p in $P_{\text{Prec}}$), while other models degrade by 3.4--6.3\%p. Examples teach \emph{how to answer} but not \emph{what to exclude}—models mimic the pattern but still blend.

Negative Constraints yield only modest gains. GPT-5 achieves the largest improvement (+3.54\%p in $P_{\text{Prec}}$), but precision still plateaus around 50\%, and other models show smaller gains or drops. Even directive-compliant models cannot suppress factually valid but off-scope content, indicating that selective exclusion is orthogonal to existing instruction-following capabilities.

\subsection{Reasoning Capability is Decoupled from Selective Exclusion}
This section examines how reasoning capability interacts with selective exclusion. Our settings progressively reduce how much a model must infer the target perspective: Base gives only a perspective hint, Oracle provides target-perspective facts, and Negative Constraints adds explicit exclusion instructions. Yet Oracle precision plateaus in the low-to-mid 40s on a 10-model average, and GPT-5, a reasoning-capable model in this near-ideal setting, moves only from 48.8 to 52.3 under Negative Constraints (Table~\ref{tab:ablation-study}). This is difficult to explain by a failure to identify the target perspective alone, suggesting that the limitation does not lie in the reach of instructions but in a capability that current alignment training does not address.

To test this interpretation, we compare GPT-5 under Oracle conditions across two reasoning-effort settings (Table~\ref{tab:reasoning-effort}). Higher reasoning effort slightly degrades $P_{\text{Prec}}$ (48.50 $\rightarrow$ 43.68), indicating that scaling reasoning does not improve—and may even slightly reduce—selective exclusion. This reinforces our central claim along the reasoning axis: inclusion and exclusion are decoupled capabilities.

\begin{table}[htbp]
    \caption{GPT-5 reasoning effort (Oracle); mean over questions. Best scores are bold.}
    \label{tab:reasoning-effort}
    \centering
    \footnotesize
    \setlength{\tabcolsep}{3pt}
    \begin{tabular}{lrrrrr}
        \toprule
        \textbf{Effort} & \textbf{$P_{\text{Prec}}$} & \textbf{$P_{\text{Rec}}$} & \textbf{PAS} & \textbf{Faithfulness} & \textbf{\#Tok} \\
        \midrule
        Low  & \textbf{48.50} & \textbf{78.66} & \textbf{55.22} & \textbf{72.98} & 274.21 \\
        High & 43.68 & 75.09 & 51.49 & 71.65 & 261.91 \\
        \bottomrule
    \end{tabular}
\end{table}

\subsection{Impact of Distractor Documents}
Oracle settings revealed that models struggle with perspective control even under ideal conditions. To examine whether this problem intensifies when retrieval introduces noise, we evaluate the Noisy setting. This setting provides documents from both target paths ($\Pi_{\mathcal{P}}^{\text{target}}$) and distractor paths ($\Pi_{\mathcal{P}}^{\text{distractor}}$) in equal numbers, simulating realistic retrieval where semantically related but perspective-misaligned content appears alongside correct information.

\begin{table}[htbp]
    \caption{Noisy results: mean over questions. Best scores per model are bold; second-best are underlined.}
    \label{tab:noisy}
    \centering
    \small
    \renewcommand{\arraystretch}{0.9}
    \resizebox{1\columnwidth}{!}{
    \begin{tabular}{lrrrrr}
        \toprule
        \textbf{Model / Setting} & \textbf{$P_{\text{Prec}}$} & \textbf{$P_{\text{Rec}}$} & \textbf{PAS} & \textbf{Faithfulness} & \textbf{\#Tok} \\
        \midrule

        \multicolumn{6}{l}{\textit{\textbf{GPT-5}}} \\
        \hspace{2mm} Base & 39.58 & 68.63 & 46.26 & 34.96 & 486.99 \\
        \hspace{2mm} Noisy & \underline{45.88} & \textbf{79.67} & \underline{54.63} & \underline{67.09} & 345.58 \\
        \hspace{2mm} Oracle & \textbf{48.75} & \underline{78.82} & \textbf{56.14} & \textbf{67.53} & 237.13 \\
        \midrule

        \multicolumn{6}{l}{\textit{\textbf{Claude-Sonnet-4.5}}} \\
        \hspace{2mm} Base & 38.08 & 63.28 & 43.84 & 43.46 & 413.76 \\
        \hspace{2mm} Noisy & \underline{38.74} & \underline{69.96} & \underline{46.51} & \textbf{69.65} & 397.90 \\
        \hspace{2mm} Oracle & \textbf{47.53} & \textbf{76.66} & \textbf{54.16} & \underline{63.57} & 376.09 \\
        \midrule

        \multicolumn{6}{l}{\textit{\textbf{Qwen3-32B}}} \\
        \hspace{2mm} Base & 39.91 & 68.02 & 46.89 & 40.34 & 275.75 \\
        \hspace{2mm} Noisy & \underline{42.86} & \underline{76.92} & \underline{52.12} & \textbf{53.36} & 256.97 \\
        \hspace{2mm} Oracle & \textbf{46.98} & \textbf{77.91} & \textbf{54.58} & \underline{51.35} & 236.44 \\
        \midrule

        \multicolumn{6}{l}{\textit{\textbf{Gemma3-27B}}} \\
        \hspace{2mm} Base & 38.53 & 65.13 & 44.48 & 48.70 & 460.77 \\
        \hspace{2mm} Noisy & \underline{44.03} & \underline{73.17} & \underline{51.04} & \underline{63.64} & 294.59 \\
        \hspace{2mm} Oracle & \textbf{46.23} & \textbf{73.63} & \textbf{52.54} & \textbf{65.90} & 281.03 \\
        \bottomrule
    \end{tabular}
        }
\end{table}

Table~\ref{tab:noisy} confirms that distractor documents systematically degrade perspective control. Traditional RAG metrics fail to capture this degradation. Answer Relevancy from \cite{ragas} (Table~\ref{tab:answer_relevancy_detailed}), which measures semantic similarity between questions and answers, shows inconsistent patterns—scores fluctuate erratically across Base, Noisy, and Oracle settings regardless of distractor presence. This failure stems from the fundamental limitation exposed in Section~\ref{sec:ambiguity}: target and distractor paths exhibit nearly identical semantic similarity to queries, making similarity-based metrics unable to distinguish perspective-aligned content from off-scope alternatives.

\begin{table}[htbp]
    \caption{Answer Relevancy. Best scores per model are bold; second-best are underlined.}
    \label{tab:answer_relevancy_detailed}
    \centering
    \footnotesize
    \setlength{\tabcolsep}{3pt}
    \renewcommand{\arraystretch}{0.95}
    \begin{tabular}{lccc}
        \toprule
        \multirow{2}{*}{\textbf{Model Name}} & \multicolumn{3}{c}{\textbf{Answer Relevancy Settings}} \\
        \cmidrule(lr){2-4}
        & \textbf{Base} & \textbf{Noisy} & \textbf{Oracle} \\
        \midrule
        Claude-Sonnet-4.5 & \underline{0.7309} & 0.7297 & \textbf{0.7650} \\
        GPT-5             & \textbf{0.6568} & 0.6363 & \underline{0.6498} \\
        Gemma3-27B-IT     & 0.6067 & \underline{0.6443} & \textbf{0.6981} \\
        Qwen3-32B         & \underline{0.7287} & \textbf{0.7294} & 0.7282 \\
        \bottomrule
    \end{tabular}
\end{table}

In contrast, PAS demonstrates monotonic ordering: Noisy settings consistently produce intermediate performance between Base and Oracle across all models. This validates a critical property of our metric: it properly quantifies perspective contamination levels rather than conflating topical relevance with constraint adherence. When documents from $\Pi_{\mathcal{P}}^{\text{distractor}}$ are introduced, PAS degrades proportionally, establishing measurable sensitivity to the exact problem we aim to capture.

The degradation from Oracle to Noisy reveals varying model robustness. GPT-5 maintains relatively stable precision, while Claude-Sonnet-4.5 shows substantial vulnerability to distractor influence. Response length patterns provide additional insight: token generation decreases progressively from Base to Noisy to Oracle. When models lack clear guidance from context, they produce longer responses—a behavior reinforced through alignment training where length correlates with perceived helpfulness~\cite{deepseek-r1}. As retrieval quality improves and uncertainty decreases, responses become more concise and focused.

\begin{figure*}[!t]
    \centering
    \includegraphics[width=1\linewidth]{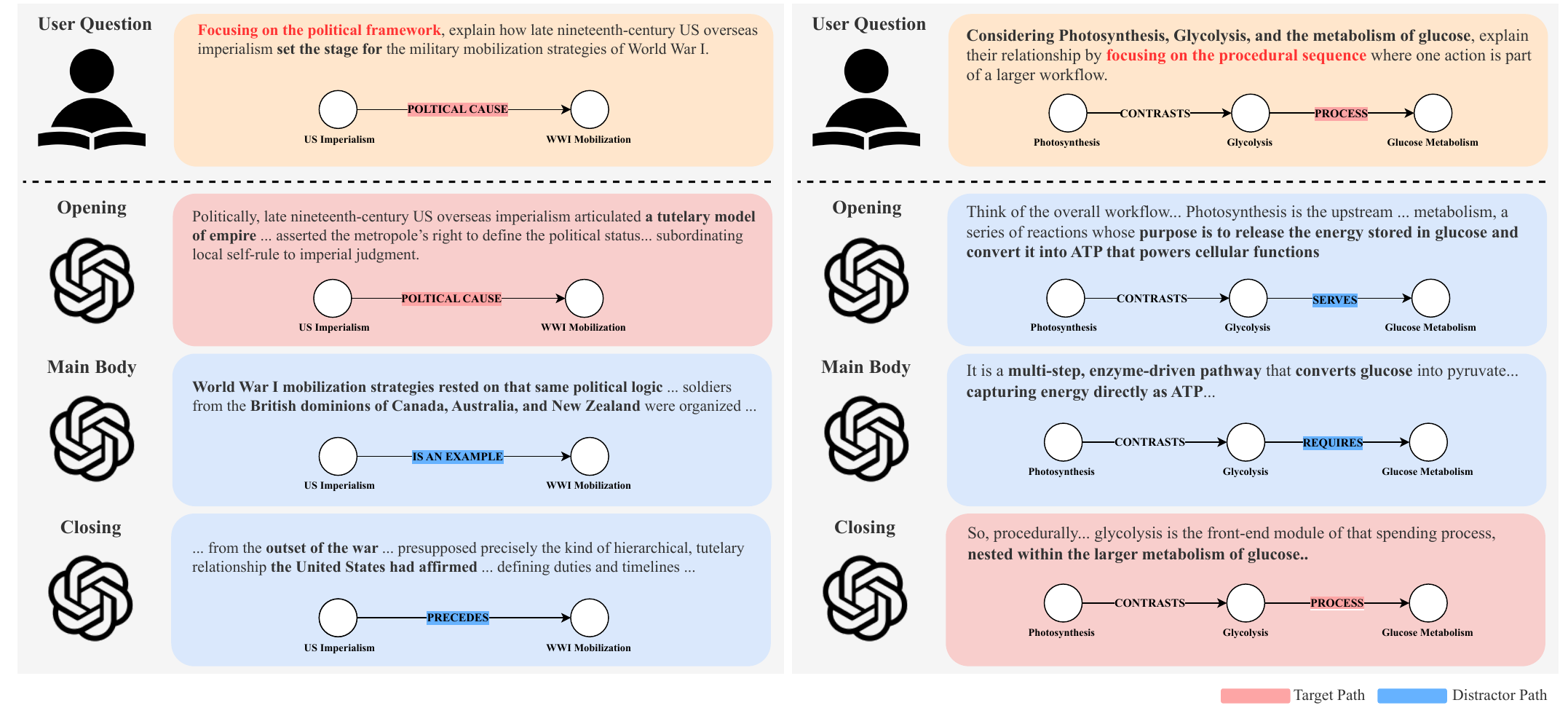}
    \caption{\textbf{Qualitative analysis of failures in selective exclusion.} The History case (left) adds unrequested entities, while the Biology case (right) adds details beyond the requested sequence. Although \textbf{factually accurate}, both responses violate \textbf{explicit constraints} by merging \textbf{multiple valid perspectives}.}
    \label{fig:qulitative}
\end{figure*}

\begin{table}[!htbp]
    \caption{Retrieval precision/recall with Noisy's document count. Best scores per model are bold; second-best are underlined.}
    \label{tab:real-retrieval}
    \centering
    \footnotesize
    \renewcommand{\arraystretch}{0.95}
    \setlength{\tabcolsep}{2.5pt}
    \begin{tabular}{lccc}
        \toprule
        \textbf{Setting} & \makecell{\textbf{Claude-}\\\textbf{Sonnet-4.5}} & \textbf{Qwen3-32B} & \textbf{Gemma3-27B} \\
        \midrule
        Base & 38.08 / 63.28 & 39.91 / 68.02 & 38.53 / 65.13 \\
        RAG BM25 & 31.0 / 59.3 & 39.4 / 70.0 & 33.7 / 64.7 \\
        RAG dense & 34.4 / 63.7 & 34.1 / 65.6 & 36.5 / 60.8 \\
        Noisy & \underline{38.74} / \underline{69.96} & \underline{42.86} / \underline{76.92} & \underline{44.03} / \underline{73.17} \\
        Oracle & \textbf{47.53} / \textbf{76.66} & \textbf{46.98} / \textbf{77.91} & \textbf{46.23} / \textbf{73.63} \\
        \bottomrule
    \end{tabular}
\end{table}

To examine whether the exclusion failure persists under real retrieval, we attached BM25 and the dense retriever BGE-large-en-v1.5 to the textbook corpus and gave each question the same number of retrieved documents as Noisy, across three models (Table~\ref{tab:real-retrieval}). Only 43.5\% (BM25) and 47.4\% (dense) of retrieved passages belonged to the requested perspective, with distractor perspectives at 32.2\% and 36.5\%. Precision remains below Noisy and Oracle for all three models, so real retrieval does not resolve the exclusion failure observed in the controlled settings.

\subsection{Explanations for Low Precision}
\label{sec:scoring-artifacts}

We examine whether low precision reflects all-or-nothing path scoring or item ambiguity. To distinguish perspective control failure from the difficulty of reproducing longer paths, we re-scored Oracle responses from GPT-5, Qwen3-32B, and Gemma3-27B with per-edge partial credit. Each path $\pi$ receives credit $c(\pi)$ equal to the fraction of its edges the response reproduces, with the same Gemini judge and partial credit applied equally to precision and recall. Precision and recall replace $|A \cap T|$ with $\sum_{\pi \in T} c(\pi)$ and $|A|$ with $\sum_{\pi \in \Pi(q)} c(\pi)$.

\begin{table}[htbp]
    \caption{Oracle recall drop and precision change, 1-hop to 3-hop (same Gemini judge).}
    \label{tab:partial-credit}
    \centering
    \small
    \resizebox{\columnwidth}{!}{
    \begin{tabular}{lcc}
        \toprule
        \textbf{Model} & \makecell{Recall drop\\all-or-nothing / partial} & \makecell{Precision change\\all-or-nothing / partial} \\
        \midrule
        GPT-5 & $-36.6$ / $-12.8$ & $-0.7$ / $-6.0$ \\
        Qwen3-32B & $-31.7$ / $-12.1$ & $+1.2$ / $+4.5$ \\
        Gemma3-27B & $-34.8$ / $-15.8$ & $\pm 0.0$ / $+3.9$ \\
        \midrule
        Average & $-34.4$ / $-13.6$ & $\approx$ flat \\
        \bottomrule
    \end{tabular}
    }
\end{table}

Table~\ref{tab:partial-credit} shows that the average 1-hop to 3-hop recall drop shrinks from 34.4 to 13.6 points under partial credit, so most of the per-hop recall loss reflects the difficulty of reproducing longer chains. Precision, by contrast, stays low under both scorings (Oracle 43--54) with no consistent trend across hops. Since the same relaxation is applied to both metrics, low precision appears unlikely to be an artifact of all-or-nothing path scoring.

To examine the effect of item ambiguity, we re-aggregated scores on items where all three annotators agreed on distractor validity (Appendix~\ref{sec:app-unanimous}). Precision does not rise noticeably on these items, suggesting that ambiguous items alone do not explain the low precision.

\subsection{Qualitative Analysis}

To illustrate how high Faithfulness can coexist with perspective misalignment, we examine 25 responses that score highly on Faithfulness---a widely used metric for whether claims are grounded in context---yet receive low PAS scores, each reviewed independently by two authors. Figure~\ref{fig:qulitative} presents two representative cases where factual accuracy and perspective adherence diverge. The red-colored parts show target paths $\Pi_{\mathcal{P}}^{\text{target}}$, while the blue-colored parts show distractor paths $\Pi_{\mathcal{P}}^{\text{distractor}}$.

In the History domain (left), the question requests a political framework connecting US imperialism to WWI mobilization. Despite high Faithfulness, the response incorporates distractor paths by introducing British dominions as exemplification and adding temporal precedence---beyond the requested political lens. In the Biology domain (right), the question asks for procedural sequence only. The response is accurate but adds functional purposes and mechanistic details from the ``Serves'' and ``Requires'' paths rather than maintaining focus on the requested ``Process'' relationship.

The key pattern is that responses include the target perspective but also incorporate content from adjacent perspectives. This is consistent with the low precision across hops under partial-credit scoring (Section~\ref{sec:scoring-artifacts}). These cases show that high Faithfulness does not ensure perspective alignment. Models incorporate semantically related content from $\Pi_{\mathcal{P}}^{\text{distractor}}$, producing comprehensive yet unfocused responses that users must filter.

\section{Related Works}
\label{sec:related}
\noindent\textbf{RAG and Instruction-Following Evaluation.}
RAG evaluation has matured around metrics for fluency, factual accuracy, and answer quality~\cite{rag_metric_survey, ragas}. In parallel, instruction-following benchmarks have progressively widened the constraint space. IFEval~\cite{ifeval} introduced 25 verifiable surface-level constraints such as length, format, and keyword inclusion. FollowBench~\cite{followbench} extended evaluation across five axes—content, situation, style, format, and example—at multiple difficulty levels, and InfoBench~\cite{infobench} measured fine-grained compliance through the Decomposed Requirements Following Ratio. ComplexBench~\cite{complexbench} brought semantic-level constraints into the taxonomy. Across both lines, evaluation predominantly measures inclusion of relevant or constraint-satisfying content. Exclusion-oriented evaluation has been explored in adjacent settings. CoDI-Eval~\cite{codieval} introduced an exclusion dimension via Toxicity Avoidance, though limited to safety-level surface cases. DeCRIM~\cite{decrim} evaluates explicit negative constraints such as ``no hashtag,'' while the exclusion constraint in EIFBENCH~\cite{eifbench} forbids specific mentions. These benchmarks do not directly measure exclusion of factually valid reasoning outside the requested perspective.

\noindent\textbf{Personalization and Persona Evaluation.}
RoleLLM~\cite{rolellm} evaluates role-level consistency through RoleBench, while PersonaGym~\cite{personagym} measures persona adherence. LaMP~\cite{lamp}, LaMP-QA~\cite{lampqa}, and PrefEval~\cite{prefeval} evaluate whether models can be steered toward a user's profile or preference. On opinion alignment, \citet{santurkar} find that steering contributes to but does not fully achieve alignment, and \citet{durmus} report that persona prompting shifts viewpoints but reinforces stereotypes. Across these axes, evaluation focuses on steering model behavior toward a target, without directly measuring exclusion of factually valid content outside the request.

\noindent\textbf{Perspective Recognition and Selective Exclusion.}
PERSPECTRA~\cite{perspectra} identifies pro and con viewpoints in a discussion: its perspectives are subjective stances, and exclusion is not involved. PerspectiveBench instead evaluates whether responses follow the requested perspective while excluding other factually valid reasoning. Our distractors are factual and as close to the query in topic as the targets (Figure~\ref{fig:sim_dist}), making them hard negatives that simple relevance scoring cannot filter. Semantic similarity metrics conflate topical relevance with constraint adherence~\cite{leveraging, sage}, masking cases where responses blend multiple perspectives that all pass factual verification. Appendix~\ref{sec:app-benchmark-comparison} provides the full comparison with per-benchmark descriptions.

\section{Conclusion}

We identified perspective misalignment—a gap where models maintain high factual accuracy yet fail to suppress content outside user-specified perspectives. PerspectiveBench reveals that state-of-the-art models achieve high recall on 1-hop questions but critically low precision—even under Oracle conditions with perfect retrieval—supporting separate evaluation of inclusion and exclusion.

Beyond benchmarks, this challenge has direct stakes in domains where perspective-specific framing matters—education requiring a single causal account, legal analysis bounded by doctrines, and interpretation of experimental results within a given theoretical framework. In such settings, factually accurate but off-scope content can make responses difficult to use for the requested task. By publicly releasing \textbf{PerspectiveBench} and its construction pipeline, we provide a foundation for evaluating perspective alignment and constructing benchmarks; extension to professional settings remains to be validated.

\section*{Limitations}

\noindent\textbf{Domain coverage.} PerspectiveBench is constructed from OpenStax textbooks in Physics, Biology, and History. These textbook questions do not span open-domain or user-facing tasks, and whether our findings hold in professional settings such as legal drafting or interpretation of experimental results has not been tested.

\noindent\textbf{Clustering sensitivity.} We have not conducted a full clustering sensitivity analysis under alternative UMAP, HDBSCAN, and top-K settings. Three annotators checked target alignment and distractor validity on the validation sample (Appendix~\ref{sec:validity}). This does not establish that the target/distractor split remains stable under different settings.

\noindent\textbf{Judge dependence.} Both PAS and Faithfulness rely on Gemini-2.5-Pro. Human validation covers PAS path judgments, not Faithfulness, for which we have no separate human validation. We re-scored PAS with GPT-5.6 Terra (Section~\ref{sec:eval-metrics}), but possible bias from benchmark construction with Gemini and bias shared across model families remain.

\noindent\textbf{Scoring granularity.} PAS evaluates each path as an atomic unit: reproducing only part of a target path receives no credit, while partial distractor paths are not identified as violations. Per-edge partial-credit results are reported in Section~\ref{sec:scoring-artifacts}. For Faithfulness, the main evaluation uses target-path documents, so factually valid distractor claims may be marked unsupported; Appendix~\ref{sec:app-expanded-gt} reports an expanded-ground-truth check.

\noindent\textbf{Human performance.} We have not measured PAS on answers written by experts. The cover-all and unanimous-item analyses do not establish a human performance level. Evaluating expert answers remains future work.

\noindent\textbf{Retrieval realism.} Noisy injects distractor documents in a controlled 1:1 ratio with targets and does not reproduce a deployed retrieval pipeline. Our BM25 and dense retrieval experiments cover three models on the textbook corpus (Table~\ref{tab:real-retrieval}); interactions between retrieval quality and perspective control in other deployed settings remain to be evaluated.

\noindent\textbf{No user study.} We have no direct user study measuring cognitive cost for this task. The literature cited in Appendix~\ref{sec:app-cost} provides background motivation.

\section*{Ethical Considerations}

\noindent\textbf{Data source and licensing.} The OpenStax source textbooks are released under a Creative Commons Attribution license that permits redistribution and adaptation with attribution. We will release the benchmark and construction pipeline under the same permissive license.

\noindent\textbf{Human validation.} Three voluntary annotators recruited from the authors' home institution performed both the 180-item inter-rater study (Appendix~\ref{sec:validity}) and the PAS judge-validation study (Section~\ref{sec:eval-metrics}). All three hold graduate-level training in NLP and machine learning and participated as part of an internal academic collaboration without compensation. Annotators were informed of the study purpose and consented to the release of aggregate inter-rater statistics; no per-annotator labels or personally identifying information are released.

\noindent\textbf{Use of LLMs.} We used Gemini-2.5-Pro for benchmark construction and evaluation of PAS and Faithfulness. We also re-scored PAS on the same responses with GPT-5.6 Terra. Prompt templates are provided in Appendix~\ref{sec:app-prompts-templates}. LLM assistance was also used for grammar polishing and revisions to the wording and structure of the paper; all technical claims, experimental design, and analyses were authored and verified by the human authors.

\noindent\textbf{Intended use and dual-use risk.} PerspectiveBench is intended for academic evaluation, not production decisions on its own. Selective exclusion could be misused to enforce one-sided narratives or suppress viewpoints. We report precision and recall alongside PAS, and Faithfulness to assess support for response claims. Exclusion should follow the user's requested scope; when that scope is not constrained, users may want broader coverage.

\section*{Acknowledgments}

This work was supported by the Technological Innovation R\&D Program (SCALEUP TIPS, Tech Investor Program for Scaleup; No. RS-2025-25465308) funded by the Ministry of SMEs and Startups (MSS, Korea). This research was supported by Global - Learning \& Academic research institution for Master's\textperiodcentered{}PhD students, and Postdocs (G-LAMP) Program of the National Research Foundation of Korea (NRF) grant funded by the Ministry of Education (No. RS-2025-25442252).

\bibliography{referene}

\appendix

\section{Why Perspective Constraints Matter}
\label{sec:app-cost}

\subsection{Constraints from the Task}
A perspective constraint is not a matter of stylistic preference. It comes from what the person receiving the answer has to do with it. Three settings make this concrete.

\begin{itemize}
\item \textbf{Education:} A student in an economic history course analyzes the fall of Rome through an economic lens. What the student has to do is explain, as a single causal account, how economic factors led to the collapse. If the model supplies the military and political causes as well, the student must separate the economic explanation from the other causes to complete the task.

\item \textbf{Law:} An attorney asked to provide an analysis under contract doctrine alone is given tort arguments alongside the contract ones. Even if both are legally sound, a draft that mixes the two cannot be used as written for the requested task.

\item \textbf{Science:} A researcher asked to interpret an experimental result thermodynamically is given a dynamical account as well. Even if both explanations are correct, mixing them does not show whether thermodynamic premises alone account for the result.
\end{itemize}

All three call for an account whose perspective is constrained out of necessity. One might argue that the user can simply pick out the parts they need, but this shifts the burden back onto the user. In multi-hop explanations especially, the perspectives are interleaved sentence by sentence, and extracting only the information belonging to the requested perspective is itself difficult.

\subsection{Cognitive Load as Background}
Contemporary LLMs have a strong tendency toward lengthy, comprehensive responses. Through instruction-following training, particularly RLHF, models learn that response length and breadth correlate with perceived helpfulness~\cite{deepseek-r1, singhal2024a}, leading to exhaustive coverage patterns~\cite{mitigating} even when users specify narrow constraints.

This verbosity can impose costs on users across multiple dimensions. Cognitively, research on information processing demonstrates that excessive information—even when factually correct—increases cognitive load and impairs decision-making efficiency~\cite{mit2025cognitive, SHAHRZADI2024100261}. Users must perform active cognitive filtering to extract relevant information from verbose responses, a task that consumes working memory resources and increases mental effort~\cite{owens2013improving}. This process, which cognitive science terms ``information triage''~\cite{zahabi2013exploring}, requires determining which information to retain and which to ignore. Unlike hallucinations, which signal clear quality failures, perspective violations present factually grounded content that appears relevant, making misalignment difficult to detect without explicit attention.

Beyond cognitive burden, longer responses require more output tokens and can increase API costs and latency.

These challenges motivate growing recognition that selective information delivery—not just comprehensive coverage—represents a critical capability for practical systems. Recent work emphasizes designing retrieval systems that align with focused user intent~\cite{10852457, Yu_2025_ICCV, zhang2025personalize}. However, evaluating such selective control requires distinguishing valid target content from valid but off-scope alternatives, a measurement challenge our benchmark directly addresses through explicit perspective partitioning and separate assessment of inclusion versus exclusion capabilities. This literature provides background motivation; we do not measure cognitive cost directly for this task.

\section{Domain Selection Rationale}
\label{sec:app-domains}

We selected three domains where single questions naturally admit multiple valid perspectives:

\begin{itemize}
\item \textbf{Physics}: Physical phenomena can be interpreted through different theoretical frameworks. For instance, energy conservation can be analyzed from mechanical or thermodynamic perspectives, where both provide valid yet distinct explanations.

\item \textbf{Biology}: Biological processes support explanations at different levels of analysis. Cellular respiration can be described through biochemical mechanisms or metabolic regulation roles, each offering a different explanatory focus.

\item \textbf{History}: Historical events arise from multidimensional causes that can be examined separately. Rome's fall can be explained through economic, military, or political factors, where each perspective provides a coherent causal narrative.
\end{itemize}

\section{Edge Score Weights}
\label{sec:app-edge-scoring}

The relevance score $\text{Score}(v_i, v_j) = S_{\text{ent}} + S_{\text{prox}} + S_{\text{sem}}$ in Section~\ref{sec:Edge} uses the following point assignments:

\noindent\textbf{Shared Entity} ($S_{\text{ent}}$). Let $E(v_i)$ denote the set of entities extracted from facts in cluster $v_i$ (from Section~\ref{sec:Extraction of Single Fact Sentences}). We assign 5 points when $|E(v_i) \cap E(v_j)| \geq 3$ and 4 points when $1 \leq |E(v_i) \cap E(v_j)| \leq 2$.

\noindent\textbf{Textual Proximity} ($S_{\text{prox}}$). Each fact $f$ is indexed by its sentence position in the source text. Facts $f_i \in C_i$ and $f_j \in C_j$ co-occur if $|pos(f_i) - pos(f_j)| \leq 5$. Let $k$ be the total co-occurring pair count; we use binary search over sorted position lists to count these pairs in $O(|C_i| \cdot \log|C_j|)$ time. We assign 3 points when $k \geq 5$, 2 points when $2 \leq k \leq 4$, and 1 point when $k = 1$.

\noindent\textbf{Probabilistic Cluster Overlap} ($S_{\text{sem}}$). When facts assigned to one cluster could plausibly belong to another (soft HDBSCAN probabilities), this indicates semantic proximity. We assign 1 point when each cluster contains at least 3 facts with soft HDBSCAN probabilities greater than 0.05 for the other cluster. 

\begin{figure*}[ht]
    \centering
    \includegraphics[width=1\linewidth]{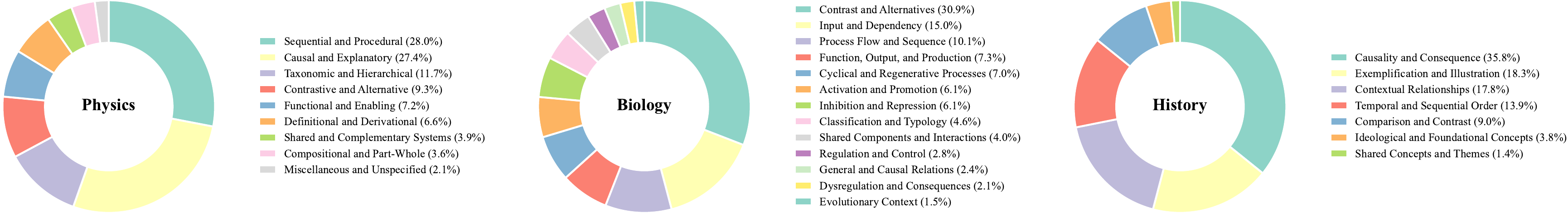}
    \caption{\textbf{Distribution of semantic perspective types across domains.} Fine-grained relationships are aggregated into high-level categories. The distributions show the range of perspective types present in each domain.}
    \label{fig:theme}
\end{figure*}

\section{Multi-Perspective QA Generation: Detailed Steps}
\label{sec:app-qa-gen}

\noindent\textbf{Step 1: Reasoning Path Construction.}
A reasoning path $\pi$ represents a sequence of nodes connected by edges in graph $G$. Formally, $\pi = (v_0 \rightarrow v_1 \rightarrow \cdots \rightarrow v_k)$ where $v_0 = v_s$ is the start node, $v_k = v_e$ is the end node, and each arrow represents an edge $(v_i, v_{i+1}) \in \mathcal{E}$ with an associated relationship label. Paths range from 1-hop (direct connection: $v_s \rightarrow v_e$) to 3-hop (multi-step: $v_s \rightarrow v_1 \rightarrow v_2 \rightarrow v_e$).

\noindent\textbf{Step 2: Perspective Definition.}
For nodes $v_s$ and $v_e$, the graph yields multiple valid paths: $\Pi(v_s \rightarrow v_e) = \{\pi_1, \pi_2, \ldots, \pi_n\}$. A perspective $\mathcal{P}_i \subseteq \Pi(v_s \rightarrow v_e)$ represents paths sharing the same explanatory focus—the type of relationship (e.g., causal, comparative, structural) connecting nodes. In Figure~\ref{fig:main}, from ``Military Decay'' to ``Fall of Western Rome,'' three distinct relationship types exist: ``External Invasion,'' ``Political Collapse,'' and ``Economic Collapse.'' Paths traversing ``Political Collapse'' or ``Economic Collapse'' share an internal explanatory focus, forming perspective $\mathcal{P}_{\text{internal}}$.

\noindent\textbf{Step 3: Perspective-Targeted Question Generation.}
For each perspective $\mathcal{P}_i$, the prompt requires an LLM to generate question $q_i$ using conceptual hints to guide models toward target paths, without using edge labels or their direct synonyms. For each generated question, we compile ground-truth documents by collecting single-fact sentences from clusters along target paths $\Pi_{\mathcal{P}}^{\text{target}}$. These documents are provided as context to models in evaluation settings that simulate ideal retrieval conditions.

\section{Distribution of Perspective Types}
\label{sec:app-theme}

To characterize the perspective types present in our benchmark, we grouped specific perspectives into semantically coherent categories using LLMs. Figure~\ref{fig:theme} shows their distributions across domains.

Physics predominantly features temporal and causal perspectives. As shown in Table~\ref{tab:domain-questions-compact}, ``Sequential and Procedural'' questions target derivation order—which principle must precede another—while ``Causal and Explanatory'' questions focus on mechanistic frameworks underlying phenomena. The same physical concepts, such as Newton's laws and momentum conservation, can be explained through either temporal precedence (which derives from which) or mechanistic causality (how one principle explains another), demonstrating that even structured domains support multiple valid reasoning paths.

Biology exhibits contrastive and dependency patterns. Questions probe functional differentiation (``Contrast and Alternatives''), prerequisite conditions (``Input and Dependency''), and workflow integration (``Process Flow and Sequence''). When examining processes like transcription versus replication, models must determine whether the question targets functional objectives, resource dependencies, or procedural sequences—each constituting a valid but distinct explanatory approach to the same biological phenomena.

History includes causal and contextual perspectives, as well as exemplification. ``Causality and Consequence'' questions span political, economic, and social dimensions. ``Exemplification and Illustration'' questions connect abstract trends to concrete events, while ``Contextual Relationships'' questions require synthesizing background frameworks. In Table~\ref{tab:domain-questions-compact}, the history questions require political causes or concrete examples of broader trends, showing how different relationships yield different explanations.

\begin{table*}[htbp]
    \caption{Representative questions from the benchmark dataset, categorized by domain and semantic perspectives.}
    \label{tab:domain-questions-compact}
    \centering
    \small
    \renewcommand{\arraystretch}{1.1} %
    \setlength{\tabcolsep}{4pt}       %
    
    \resizebox{\textwidth}{!}{
    \begin{tabular}{l l >{\small}p{11.7cm}} %
        \toprule
        \textbf{Domain} & \textbf{Category} & \textbf{Representative Question} \\
        \midrule
        
        \multirow{6}{*}{\textbf{Physics}} 
        & \multirow{2}{*}{\textbf{Sequential and Procedural}} 
        & Describe the logical sequence connecting Newton's second law of motion and the conservation of momentum, focusing on which concept must be established first to derive the other. \\
        \cmidrule{2-3}
        
        & \multirow{2}{*}{\textbf{Causal and Explanatory}} 
        & Focusing on the underlying mechanism, describe how the vector summation of forces provides a framework for understanding an object's state of constant velocity. \\
        \cmidrule{2-3}
        
        & \multirow{2}{*}{\textbf{Taxonomic and Hierarchical}} 
        & Explain the connection between mechanical energy, conservation of momentum, Newton's second law, and Newton's third law, focusing on the hierarchical relationship. \\
        \midrule
        
        \multirow{6}{*}{\textbf{Biology}} 
        & \multirow{2}{*}{\textbf{Contrast and Alternatives}} 
        & Focusing on their fundamental objectives and scope, explain what makes the process of gene transcription distinct from DNA replication. \\
        \cmidrule{2-3}
        
        & \multirow{2}{*}{\textbf{Input and Dependency}} 
        & Considering chemiosmosis, the evolution of aerobic respiration, and fermentation, explain their relationship by focusing on the essential resources one process provides for another to function. \\
        \cmidrule{2-3}
        
        & \multirow{2}{*}{\textbf{Process Flow and Sequence}} 
        & Considering Photosynthesis, Glycolysis, and the metabolism of glucose, explain their relationship by focusing on the procedural sequence where one action is part of a larger workflow. \\
        \midrule
        
        \multirow{6}{*}{\textbf{History}} 
        & \multirow{2}{*}{\textbf{Causality and Consequence}} 
        & Focusing on the political chain of events, explain the connection between the Balkan crisis, German foreign policy, and the rise of socialism. \\
        \cmidrule{2-3}
        
        & \multirow{2}{*}{\textbf{Exemplification and Illustration}} 
        & Focusing on how the Balkan crisis serves as a specific illustration of broader trends, explain the connection between it, German foreign policy, and the rise of socialism. \\
        \cmidrule{2-3}
        
        & \multirow{2}{*}{\textbf{Contextual Relationships}} 
        & Describe the broader historical framework established by British mercantilist policies within which the mobilization of armies during World War I took place. \\
        \bottomrule
    \end{tabular}
    }
\end{table*}
\section{Benchmark Construction Validity}
\label{sec:validity}
Our pipeline combines deterministic algorithms with LLM generation, so the deterministic component alone cannot certify the validity of all outputs. We therefore conduct a human audit on a stratified validation sample of 180 QA items (three domains × three hop levels × 20 items per stratum). This size yields a margin of error of $\pm 7.3$\%p at the 95\% confidence level for binary proportion estimates, with equal allocation across strata so that overall estimates are computed directly over the full sample without post-hoc weighting. This audit checks the final target/distractor split on the validation sample, rather than treating clustering output alone as evidence of validity.

Three annotators with graduate-level training in the relevant domains independently labeled all 180 items, following a 20-item pilot calibration round in which agreement of $\geq 0.70$ was required before proceeding. Each item is judged on two binary criteria: (i) \emph{target alignment}---whether the question and its target paths $\Pi_{\mathcal{P}}^{\text{target}}$ are consistent with the specified perspective---and (ii) \emph{distractor validity}---whether each distractor path is factually valid yet distinguishable from the target. Final gold labels are assigned by majority vote across the three raters, with no inter-rater discussion to preserve full independence.

To account for the well-known deflation of Cohen/Fleiss' $\kappa$ under skewed prevalence~\cite{feinstein1990high}, we report three complementary statistics: Fleiss' $\kappa$~\cite{fleiss1971measuring} as the chance-corrected baseline, the prevalence-adjusted bias-adjusted $\kappa$ (PABAK)~\cite{byrt1993bias} as a prevalence-robust counterpart, and raw Percent Agreement as a direct observation. All metrics are computed over the full 180-item sample.

Target alignment achieves Fleiss' $\kappa$ = 0.534 (moderate~\cite{landis1977measurement}). Distractor validity achieves \textbf{PABAK = 0.638 (substantial)} together with high raw agreement (\textbf{81.9\% pairwise, 72.8\% all-rater}); the lower Fleiss' $\kappa$ of 0.421 is consistent with $\kappa$ deflation under high positive prevalence (80.6\%)~\cite{feinstein1990high}. Item-level error analysis on the disagreement subset further attributes residual disagreement primarily to borderline cases where a distractor path is topically adjacent to the target perspective---precisely the hard-negative regime our benchmark is designed to expose (Section~\ref{sec:ambiguity}). The remaining disagreement may therefore reflect both benchmark difficulty and annotation noise; restricting to the items where all three annotators agreed on distractor validity leaves precision essentially unchanged (Appendix~\ref{sec:app-unanimous}).

\section{Experimental Setup Details}
\label{sec:app-exp-setup}

\paragraph{Benchmark Construction.}
We used Gemini-2.5-Pro during benchmark construction (Section~\ref{section:construction}), including atomic fact extraction, cluster summarization, edge generation, and QA generation.

\paragraph{Decoding Parameters.}
For all evaluated models, temperature is set to 0.0 in the generation settings.

\paragraph{Evaluated Models.}
We evaluate two groups of frontier LLMs. The first group consists of state-of-the-art proprietary models: GPT-5, GPT-4.1, Claude-Sonnet-4.5, and Claude-Sonnet-4. The second group evaluates performance across model scales: the Qwen3 series (32B, 14B, 8B) and the Gemma3 series (27B, 12B, 4B).

\paragraph{LLM-as-a-Judge.}
For checklist-based evaluation of PAS and Faithfulness, we employ Gemini-2.5-Pro as the judge model, selected for its strong reasoning capabilities. We also re-scored PAS on the same responses from Claude-Sonnet-4.5, Qwen3-32B, and Gemma3-27B with GPT-5.6 Terra, with the scoring prompt and output format unchanged (Appendix~\ref{sec:app-judge-robustness}).

\paragraph{Oracle Setting Details.}
In the Oracle setting, at least two documents are provided per question, collectively containing all necessary information for the target perspective. Documents are the single-fact sentences drawn from clusters along the target reasoning paths $\Pi_{\mathcal{P}}^{\text{target}}$ (Section~\ref{sec:Perspective}).

\paragraph{Real Retrieval Setting.}
We attached BM25 (Okapi, $k_1=1.5$, $b=0.75$) and the dense retriever BGE-large-en-v1.5 (\texttt{BAAI/bge-large-en-v1.5}) to the OpenStax chunks described in Section~\ref{sec:Source Dataset}. For dense retrieval, we add \texttt{Represent this sentence for searching relevant passages: } before each question. The number of retrieved documents is $2 \times$ the number of target facts, matching Noisy. Answer generation uses the same system prompt and QUESTION / CONTEXT format as Oracle and Noisy; scoring follows the main evaluation protocol.

\paragraph{Reproducibility.}
To support transparency and enable community research, we publicly release our benchmark construction pipeline, datasets, and all prompt templates used in experiments on GitHub.\footnote{URL anonymized for review.}

\section{Prompt Templates}
\label{sec:app-prompts-templates}

We release all prompt templates used by PerspectiveBench together with our code, organized along the same pipeline as the main text: benchmark construction (Section~\ref{section:construction}), answer generation under Base and Oracle settings (Appendix~\ref{sec:app-exp-setup}), LLM-as-a-Judge evaluation (Section~\ref{sec:eval-metrics}), and the prompting-strategy variants compared in our analysis (Section~\ref{sec:prompting}).

\subsection{Benchmark Construction Prompts}
\begin{figure*}[ht]
    \centering
    \includegraphics[width=0.95\linewidth]{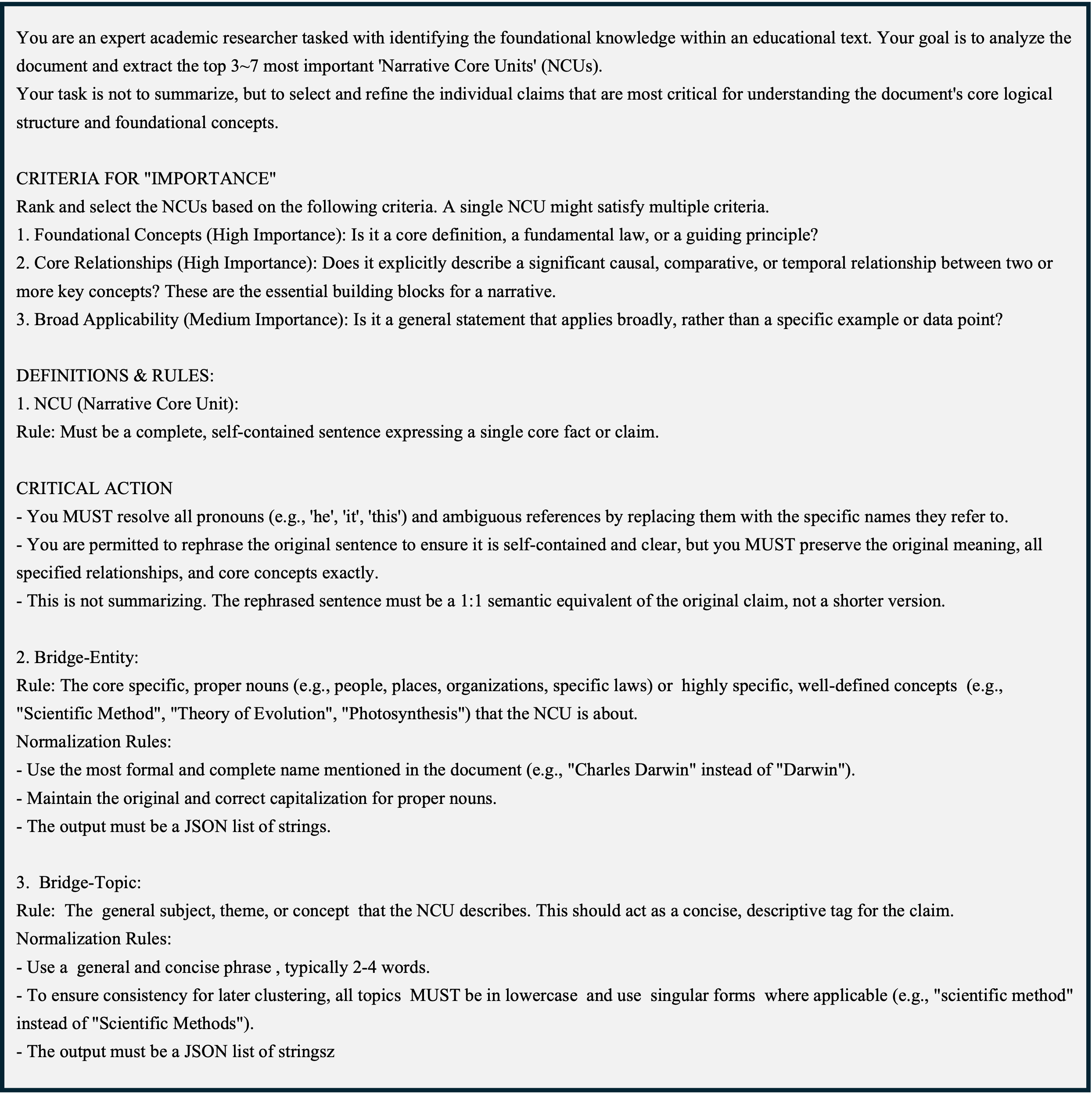}
    \caption{\textbf{Prompt template for extraction of single-fact sentences and entities (Section~\ref{sec:Extraction of Single Fact Sentences}).} The LLM decomposes each text chunk $T$ into single-fact sentences $\mathcal{F} = \{f_1, \ldots, f_n\}$ where each $f_i$ contains exactly one discrete piece of information, and pairs every $f_i$ with its key entities $e_i$. The prompt requires pronoun resolution and 1:1 semantic preservation so that each $f_i$ remains self-contained when later (i) grouped into semantic clusters in Section~\ref{sec:Semantic Fact Clustering} and (ii) served as a reference document during evaluation. The two outputs feed the next two construction stages in complementary ways: the single-fact sentences seed the density-based clustering of Section~\ref{sec:Semantic Fact Clustering}, while the entities $e_i$ contribute the shared-entity signal $S_{\text{ent}}$ used by the edge-scoring heuristic in Section~\ref{sec:Edge}.}
    \label{fig:prompt-step1}
\end{figure*}

\begin{figure*}[ht]
    \centering
    \includegraphics[width=0.95\linewidth]{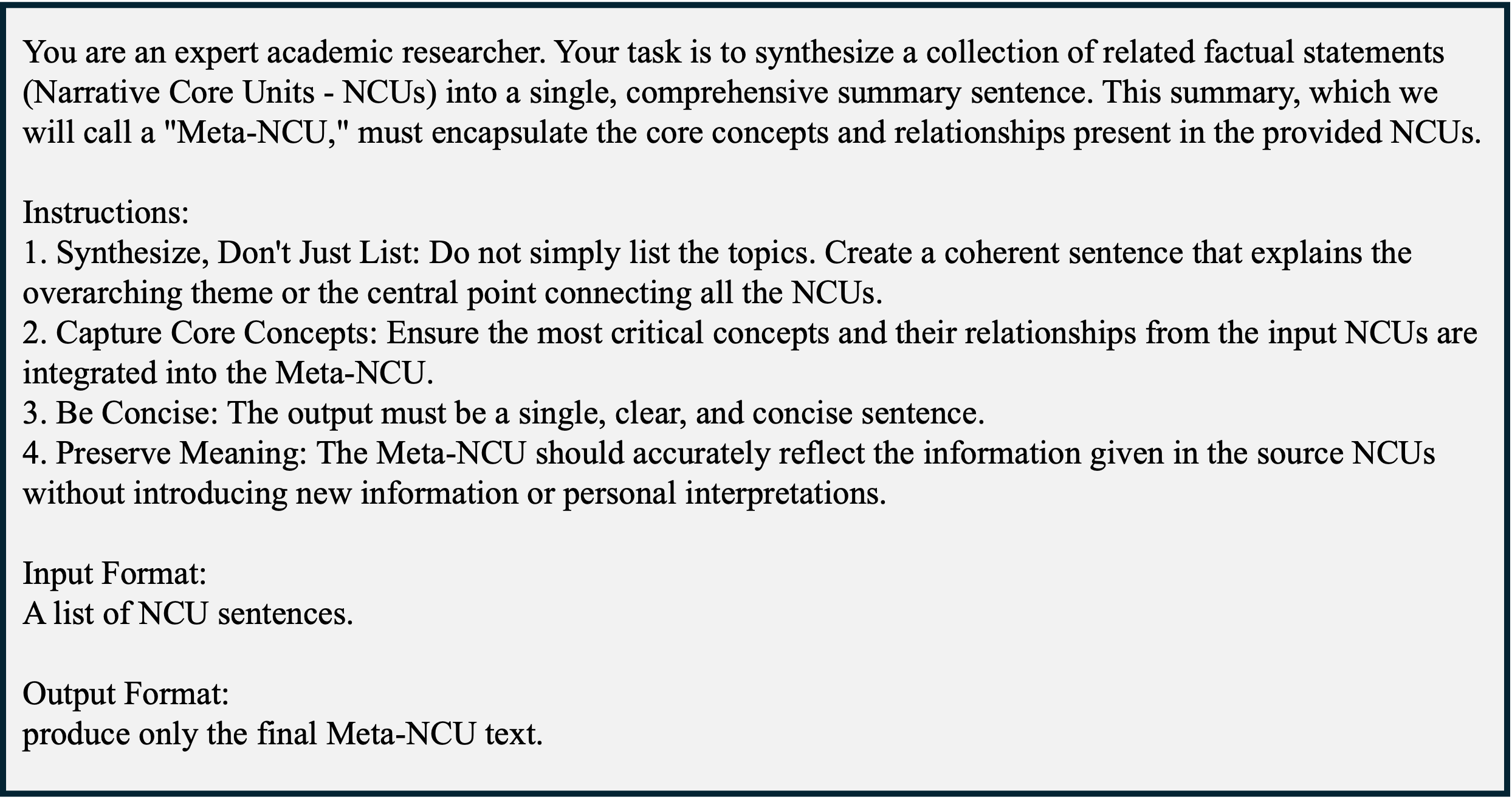}
    \caption{\textbf{Prompt template for representative summary generation (Section~\ref{sec:Semantic Fact Clustering}).} After single-fact sentences in $\mathcal{F}$ are grouped into cohesive clusters $\{C_1, \ldots, C_k\}$, each cluster $C_j$ needs an explicit concept label to function as a graph node. The LLM synthesizes the facts in $C_j$ into a single representative summary $s_j$ that names the shared explanatory focus---for example, condensing facts about inflation and civil wars into ``Internal Crisis'' as in Section~\ref{sec:Semantic Fact Clustering}. The instructions require synthesis rather than enumeration and forbid information not present in the input, so that $s_j$ faithfully labels $C_j$. Each $(C_j, s_j)$ pair becomes a semantic node $v_j \in \mathcal{V}$, on top of which edges and perspectives are constructed in Sections~\ref{sec:Edge} and~\ref{sec:Perspective}.}
    \label{fig:prompt-step2}
\end{figure*}

\begin{figure*}[ht]
    \centering
    \includegraphics[width=0.95\linewidth]{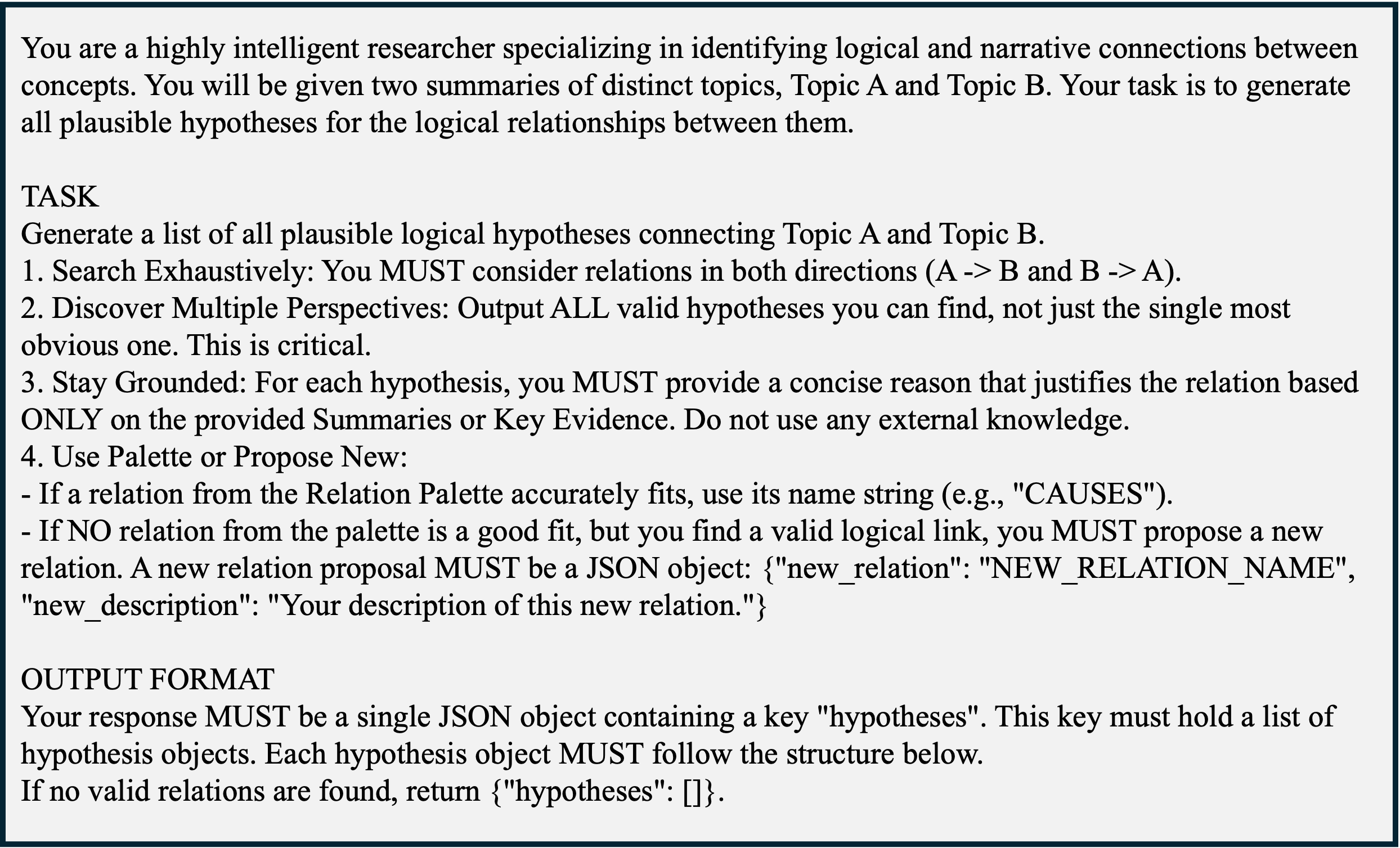}
    \caption{\textbf{Prompt template for edge generation (Section~\ref{sec:Edge}).} For each top-scoring node pair $(v_i, v_j)$ that passes the heuristic relevance score $\text{Score}(v_i, v_j) = S_{\text{ent}} + S_{\text{prox}} + S_{\text{sem}}$, the LLM receives the two representative summaries with supporting facts and emits one directed labeled edge per explanatory relationship it identifies. The prompt requires (i) bidirectional search ($v_i \rightarrow v_j$ and $v_j \rightarrow v_i$), (ii) exhaustive enumeration rather than a single best label, and (iii) grounding each relation in the provided evidence, so a node pair may carry multiple edges with distinct labels in $\mathcal{E}$. This is what lets a single endpoint pair $(v_s, v_e)$ later admit multiple valid reasoning paths $\Pi(v_s \rightarrow v_e)$---e.g., ``External Invasion,'' ``Political Collapse,'' and ``Economic Collapse'' from ``Military Decay'' to ``Fall of Western Rome'' in Figure~\ref{fig:main}.}
    \label{fig:prompt-step3}
\end{figure*}

\begin{figure*}[ht]
    \centering
    \includegraphics[width=0.95\linewidth]{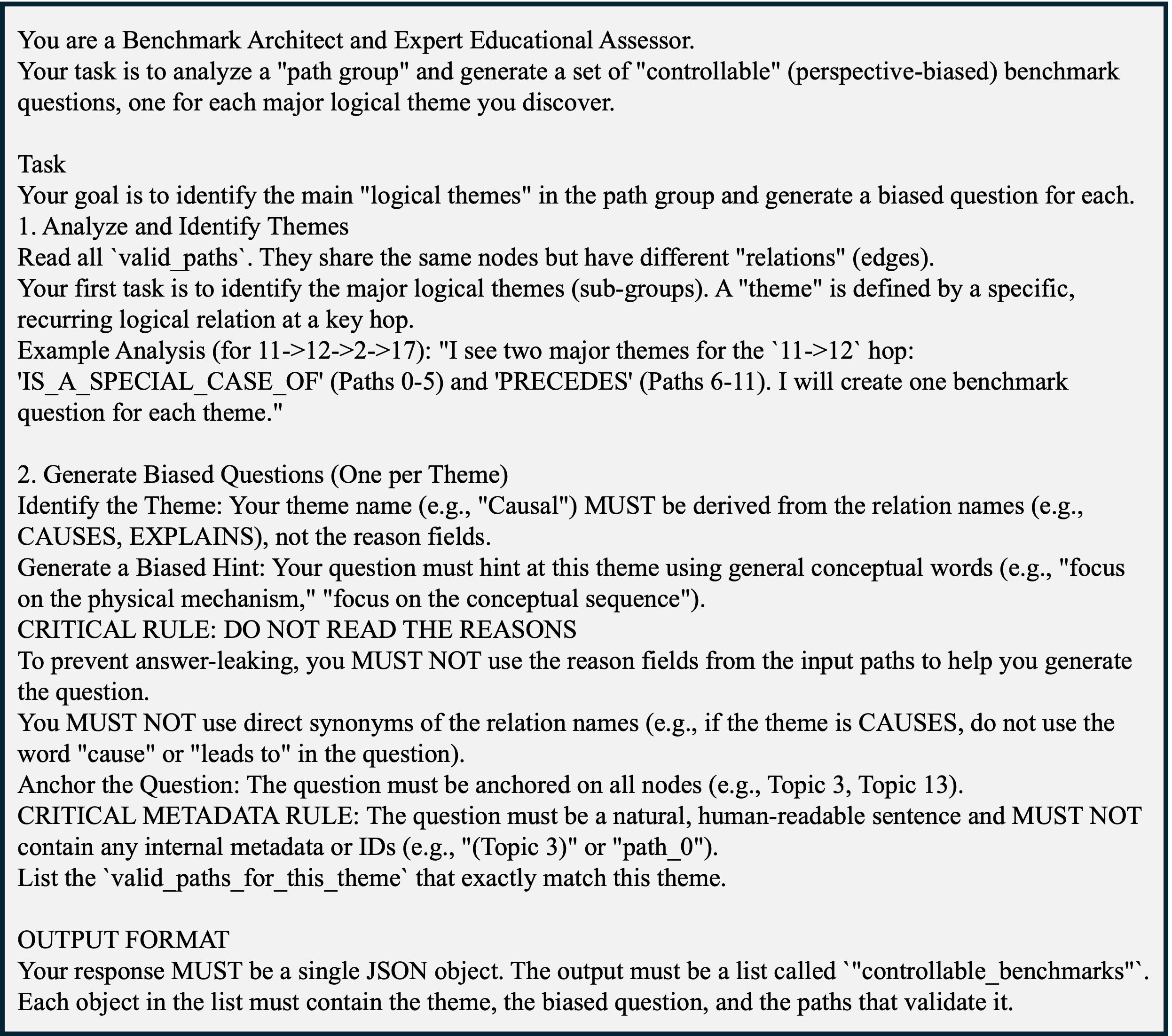}
    \caption{\textbf{Prompt template for perspective-targeted question generation (Section~\ref{sec:Perspective}).} Given the valid paths $\Pi(v_s \rightarrow v_e)$ between an endpoint pair, the LLM groups paths that share the same explanatory focus into a perspective $\mathcal{P}_i \subseteq \Pi(v_s \rightarrow v_e)$ (e.g., causal, comparative, structural), generates one question $q_i$ per perspective, and records the subset $\Pi_{\mathcal{P}}^{\text{target}}$ of paths that validate it. The prompt forbids using the \texttt{reason} fields of paths and direct synonyms of relation labels. It requires general conceptual hints to indicate the target perspective. Full step-by-step formalism is in Appendix~\ref{sec:app-qa-gen}.}
    \label{fig:prompt-step4}
\end{figure*}

Figures~\ref{fig:prompt-step1}--\ref{fig:prompt-step4} correspond, one-to-one, to the four LLM-driven construction stages and trace how raw text chunks $T$ are progressively lifted into the knowledge graph $G=(\mathcal{V}, \mathcal{E})$ and finally into perspective-controlled QA items: (i) extraction of single-fact sentences with their entities (Section~\ref{sec:Extraction of Single Fact Sentences}), (ii) representative summary generation that turns clusters into semantic nodes $\mathcal{V}$ (Section~\ref{sec:Semantic Fact Clustering}), (iii) edge generation that enumerates directed explanatory relations between nodes (Section~\ref{sec:Edge}), and (iv) perspective-targeted question generation over reasoning paths $\Pi(v_s \rightarrow v_e)$ (Section~\ref{sec:Perspective}).

\subsection{Answer Generation Prompt}
\begin{figure*}[ht]
    \centering
    \includegraphics[width=0.95\linewidth]{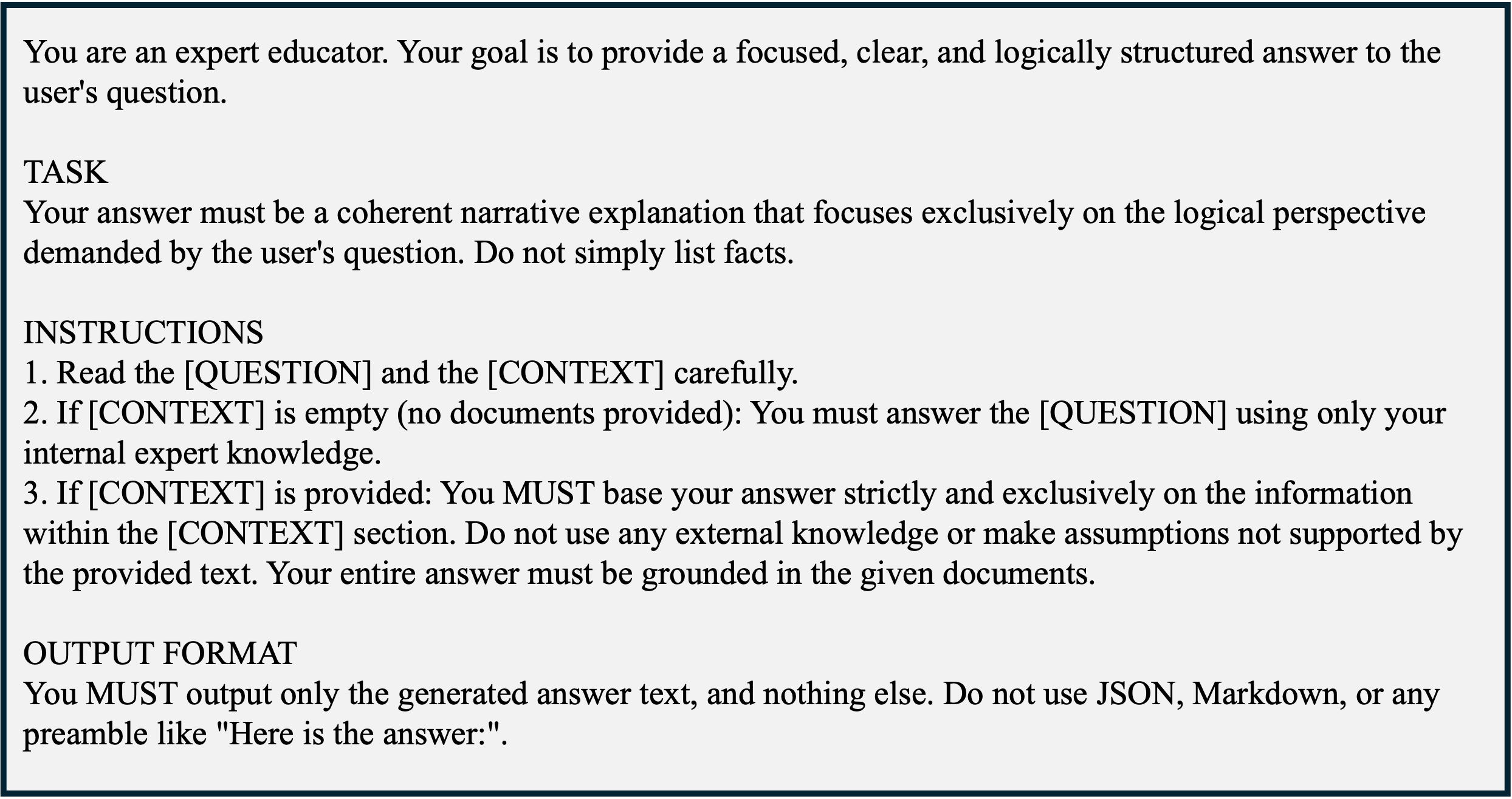}
    \caption{\textbf{Answer generation prompt for Base and Oracle settings (Appendix~\ref{sec:app-exp-setup}).} The evaluated model receives a perspective-targeted question $q_i$ produced by the Step~4 prompt (Figure~\ref{fig:prompt-step4}) and is instructed to return a coherent narrative explanation focused exclusively on the requested perspective, not an enumerated list of facts. The template branches on the \texttt{[CONTEXT]} slot: when empty, the model answers from parametric knowledge alone (Base setting); when populated with the ground-truth single-fact sentences along target paths $\Pi_{\mathcal{P}}^{\text{target}}$, the model must ground its answer strictly in those reference documents (Oracle setting).}
    \label{fig:prompt-answer}
\end{figure*}

The same answer-generation prompt (Figure~\ref{fig:prompt-answer}) is used for every evaluated model under both the Base and Oracle settings (Appendix~\ref{sec:app-exp-setup}). The context is empty in Base and contains target-path documents in Oracle.

\subsection{LLM-as-a-Judge Prompts}
\begin{figure*}[ht]
    \centering
    \includegraphics[width=0.95\linewidth]{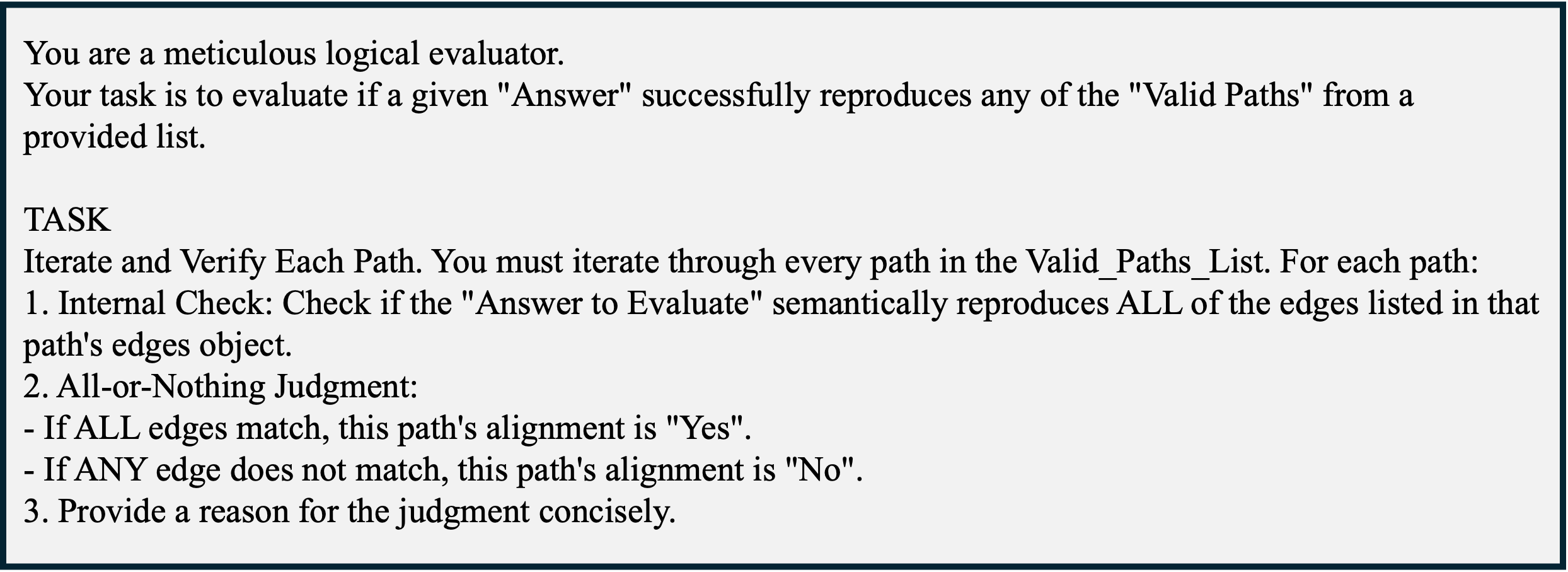}
    \caption{\textbf{LLM-as-a-Judge prompt for the Perspective Alignment Score (Section~\ref{sec:eval-metrics}).} Following the CheckEval-style checklist formulation~\cite{checkeval}, the judge does not score a response holistically. It iterates through every reasoning path $\pi \in \Pi(q)$ in the supplied \texttt{Valid\_Paths\_List} and, for each path, returns an all-or-nothing judgment of whether the response semantically reproduces \emph{all} of its constituent edges. The resulting identified set $A = \{\pi \in \Pi(q) : \text{Judge}(\pi, \text{response}) = \text{present}\}$ yields perspective precision $P_{\text{Prec}} = |A \cap T|/|A|$, recall $P_{\text{Rec}} = |A \cap T|/|T|$, and their harmonic mean PAS. Treating each path as an atomic unit is a deliberate granularity choice; sub-path violations are discussed in the Limitations.}
    \label{fig:prompt-pas}
\end{figure*}

\begin{figure*}[ht]
    \centering
    \includegraphics[width=0.95\linewidth]{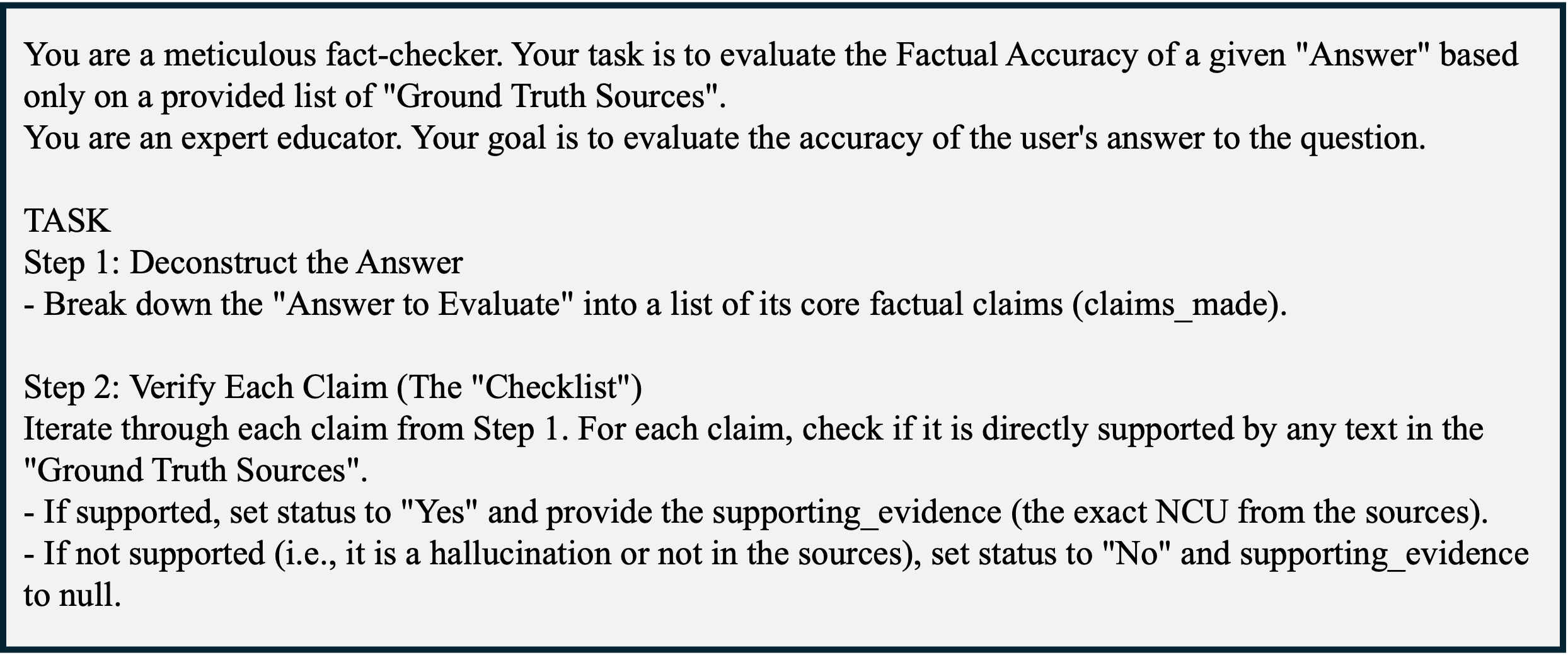}
    \caption{\textbf{LLM-as-a-Judge prompt for Faithfulness (Section~\ref{sec:eval-metrics}).} Following the standard claim-level verification protocol~\cite{ragas}, the prompt is organized into two stages. \textbf{Stage~1 (Deconstruct):} the judge breaks the response into a list of discrete factual claims (\texttt{claims\_made}). \textbf{Stage~2 (Verify):} each claim is checked against the \texttt{Ground\_Truth\_Sources}---the ground-truth single-fact sentences drawn from target paths $\Pi_{\mathcal{P}}^{\text{target}}$---and marked Yes (with the exact supporting single-fact sentence) or No (unsupported). Faithfulness is the proportion of supported claims. In the Oracle setting these same documents are exposed to the model during generation; in the Base setting they are used only post-hoc to verify the claims in the response (Section~\ref{sec:eval-metrics}).}
    \label{fig:prompt-faithfulness}
\end{figure*}

Figures~\ref{fig:prompt-pas} and~\ref{fig:prompt-faithfulness} implement the PAS and Faithfulness metrics defined in Section~\ref{sec:eval-metrics}.

\subsection{Prompting Strategy Variants}
\label{sec:app-prompts}
For the analysis in Section~\ref{sec:prompting}, we compare three prompting strategies layered on top of the answer-generation prompt of Figure~\ref{fig:prompt-answer}:
\begin{itemize}
\item \textbf{Standard Zero-shot:} basic instruction to ``focus on the requested perspective.''
\item \textbf{One-Shot Demonstration:} adds one exemplary response that perfectly adheres to perspective constraints, testing whether models can mimic the pattern through in-context learning.
\item \textbf{Negative Constraints:} adds explicit prohibitions to ``exclude information and entities from other perspectives,'' directly targeting comprehensive-response tendencies and the hard-negative distractor paths in $\Pi_{\mathcal{P}}^{\text{distractor}}$.
\end{itemize}

\section{Comparison with Existing Benchmarks}
\label{sec:app-benchmark-comparison}

Table~\ref{tab:benchmark-comparison} shows a comparison of existing benchmarks and PerspectiveBench in terms of evaluation focus, scoring, and exclusion targets. The distinction is the nature of the excluded content: our distractors are factual and as close to the query in topic as the targets (Figure~\ref{fig:sim_dist}).

\begin{table*}[t]
    \caption{Comparison with existing benchmarks. N/A indicates that no separate exclusion target is specified in this comparison.}
    \label{tab:benchmark-comparison}
    \centering
    \small
    \setlength{\tabcolsep}{4pt}
    \renewcommand{\arraystretch}{1.08}
    \begin{tabular}{>{\raggedright\arraybackslash}p{0.19\textwidth}>{\raggedright\arraybackslash}p{0.27\textwidth}>{\raggedright\arraybackslash}p{0.22\textwidth}>{\raggedright\arraybackslash}p{0.24\textwidth}}
        \toprule
        \textbf{Benchmark} & \textbf{What it measures} & \textbf{Violation detected by} & \textbf{Nature of excluded content} \\
        \midrule
        IFEval~\cite{ifeval} & Verifiable format and keyword constraints & Rules (code) & Keyword exclusion \\
        FollowBench~\cite{followbench} & Multi-level constraint following & Rules + LLM & Mostly N/A \\
        InfoBench~\cite{infobench} & Decomposed atomic requirements & LLM judge & N/A \\
        ComplexBench~\cite{complexbench} & Composed multi-constraint following & Rules + LLM & Partial (semantic) \\
        EIFBENCH~\cite{eifbench} & Complex multi-task following with exclusion & LLM judge & Specific mentions \\
        RealInstruct (DeCRIM)~\cite{decrim} & Real user instructions with multiple constraints & Human + LLM & Explicit negative constraints \\
        CoDI-Eval~\cite{codieval} & Instruction constraints and toxicity avoidance & Rules + models; Perspective API for toxicity & Keywords; toxicity \\
        LaMP~\cite{lamp} & Profile-conditioned answers & Automatic vs. reference & No exclusion notion \\
        LaMP-QA~\cite{lampqa} & Personalized long-form QA & LLM judge & No exclusion notion \\
        PrefEval~\cite{prefeval} & Adherence to stated user preferences & LLM judge & A preference, not fact \\
        RoleLLM~\cite{rolellm} & Persona speech and knowledge & Similarity + LLM & No exclusion notion \\
        PersonaGym~\cite{personagym} & Persona-agent behavioral consistency & LLM judge & No exclusion notion \\
        PERSPECTRA~\cite{perspectra} & Identifying pro/con viewpoints & Gold labels from debate graphs & Subjective stances; no exclusion notion \\
        \textbf{PerspectiveBench} & Answering within the requested perspective only & KG path checklist + LLM judge & Factual and topically as close as the target \\
        \bottomrule
    \end{tabular}
\end{table*}

\paragraph{Instruction Following.}
IFEval evaluates verifiable format and keyword constraints, including keyword exclusion. FollowBench evaluates constraints across content, situation, style, format, and example. InfoBench measures compliance through decomposed atomic requirements. ComplexBench evaluates composed multi-constraint following, including semantic constraints. EIFBENCH includes an exclusion constraint that forbids specific mentions rather than constraining an entire reasoning scope. RealInstruct, introduced with DeCRIM, evaluates real user instructions with explicit negative constraints such as ``no hashtag.'' CoDI-Eval evaluates keyword constraints and toxicity avoidance, using rules and models rather than a checklist of factual reasoning paths.

\paragraph{Personalization and Persona.}
LaMP evaluates profile-conditioned answers, while LaMP-QA evaluates personalized long-form QA. PrefEval measures adherence to stated user preferences. RoleLLM evaluates persona speech and knowledge through RoleBench, and PersonaGym measures persona-agent behavioral consistency. These benchmarks concern steering toward a profile, preference, or persona.

\paragraph{Perspective Recognition and Exclusion.}
PERSPECTRA identifies pro and con viewpoints in discourse; these perspectives are subjective stances. PerspectiveBench instead measures whether models reproduce the requested reasoning paths while excluding other factually valid paths. Topical relevance alone cannot distinguish these target and distractor paths.

\section{Judge Robustness}
\label{sec:app-judge-robustness}

We re-scored PAS on the same responses from Claude-Sonnet-4.5, Qwen3-32B, and Gemma3-27B in Base and Oracle with a judge from a different model family, GPT-5.6 Terra. We kept the scoring prompt and output format unchanged and replaced only the judge model. Each decision indicates whether the response reproduces a reasoning path.

\begin{table}[htbp]
    \caption{Agreement at the path-level Yes/No unit.}
    \label{tab:judge-agreement}
    \centering
    \footnotesize
    \setlength{\tabcolsep}{2.5pt}
    \renewcommand{\arraystretch}{1.15}
    \begin{tabular}{@{}>{\raggedright\arraybackslash}p{0.29\columnwidth}>{\raggedright\arraybackslash}p{0.28\columnwidth}>{\centering\arraybackslash}p{0.19\columnwidth}>{\centering\arraybackslash}p{0.17\columnwidth}@{}}
        \toprule
        \textbf{Comparison} & \textbf{Decision unit} & \textbf{Agreement} & \textbf{Cohen's $\kappa$} \\
        \midrule
        Human / Gemini & 220 path-level decisions & 82.7\% & 0.64 \\
        GPT-5.6 Terra / Gemini & Path-level decisions & 82.3\% & 0.645 \\
        \bottomrule
    \end{tabular}
\end{table}

\begin{table}[htbp]
    \caption{PAS evaluation under two judges. Each value pair reports Gemini / GPT-5.6 Terra. GPT-5.6 Terra re-scores the same responses.}
    \label{tab:judge-model-scores}
    \centering
    \footnotesize
    \setlength{\tabcolsep}{2.5pt}
    \renewcommand{\arraystretch}{1.15}
    \begin{tabular}{@{}lccc@{}}
        \toprule
        \textbf{Model} & \textbf{Precision} & \textbf{Recall} & \textbf{PAS} \\
        \midrule
        \multicolumn{4}{@{}l}{\textit{Base}} \\
        Claude-Sonnet-4.5 & 38.08 / 36.0 & 63.28 / 63.9 & 43.84 / 42.8 \\
        Qwen3-32B & 39.91 / 35.3 & 68.02 / 62.3 & 46.89 / 41.0 \\
        Gemma3-27B & 38.53 / 36.7 & 65.13 / 65.6 & 44.48 / 43.0 \\
        \addlinespace[2pt]
        \multicolumn{4}{@{}l}{\textit{Oracle}} \\
        Claude-Sonnet-4.5 & 47.53 / 44.6 & 76.66 / 71.8 & 54.16 / 51.4 \\
        Qwen3-32B & 46.98 / 43.2 & 77.91 / 74.9 & 54.58 / 51.1 \\
        Gemma3-27B & 46.23 / 42.9 & 73.63 / 68.9 & 52.54 / 48.3 \\
        \bottomrule
    \end{tabular}
\end{table}

Agreement between the judges is close to human agreement with Gemini (Table~\ref{tab:judge-agreement}). Recall remains higher than precision, and Oracle PAS remains higher than Base for every model under both judges. Model order can change across judges, and residual judge bias may remain.

\section{Faithfulness under an Expanded Ground Truth}
\label{sec:app-expanded-gt}

We re-scored Faithfulness for four models with a ground truth that includes distractor facts. The expanded ground truth contains all sentences in the question's graph, including every distractor-perspective sentence (on average 15 sentences per question, compared with 4 target-path sentences), so factually correct distractor content is no longer penalized for not appearing in the target documents. This is an additional check of the Base--Oracle gap; the main evaluation uses the target-path ground truth defined in Section~\ref{sec:eval-metrics}.

\begin{table}[htbp]
    \caption{Oracle $-$ Base Faithfulness gap. Target GT uses target-path sentences; full GT includes all sentences in the question's graph. Change is full GT $-$ target GT.}
    \label{tab:expanded-gt}
    \centering
    \footnotesize
    \setlength{\tabcolsep}{2.5pt}
    \renewcommand{\arraystretch}{1.15}
    \begin{tabular}{@{}lccc@{}}
        \toprule
        \textbf{Model} & \shortstack{\textbf{Gap}\\\textbf{(target GT)}} & \shortstack{\textbf{Gap}\\\textbf{(full GT)}} & \textbf{Change} \\
        \midrule
        GPT-5 & 32.6 & 32.1 & $-0.5$ \\
        Claude-Sonnet-4.5 & 20.1 & 19.2 & $-0.9$ \\
        Gemma3-27B & 17.2 & 15.4 & $-1.8$ \\
        Qwen3-32B & 11.0 & 15.1 & $+4.1$ \\
        \bottomrule
    \end{tabular}
\end{table}

For these models, the gap is largely preserved under the expanded ground truth (Table~\ref{tab:expanded-gt}), suggesting that the Base--Oracle difference is not primarily explained by penalizing factually correct distractor content.

\section{Unanimous-Item Analysis}
\label{sec:app-unanimous}

We re-aggregated the scores of GPT-5, Claude-Sonnet-4.5, Qwen3-32B, and Gemma3-27B on items where all three annotators agreed on distractor validity, as defined in Appendix~\ref{sec:validity}. The scoring is unchanged; this analysis checks whether low precision is driven by ambiguous perspective boundaries.

\begin{table}[htbp]
    \caption{Precision on the original evaluation / unanimous items. Right values report scores on items where all three annotators agreed on distractor validity under the same scoring.}
    \label{tab:unanimous-precision}
    \centering
    \footnotesize
    \setlength{\tabcolsep}{2.5pt}
    \renewcommand{\arraystretch}{1.15}
    \begin{tabular}{@{}lcc@{}}
        \toprule
        \textbf{Model} & \textbf{Base} & \textbf{Oracle} \\
        \midrule
        GPT-5 & 39.58 / 38.2 & 48.75 / 43.2 \\
        Claude-Sonnet-4.5 & 38.08 / 38.2 & 47.53 / 48.9 \\
        Qwen3-32B & 39.91 / 41.2 & 46.98 / 43.7 \\
        Gemma3-27B & 38.53 / 41.2 & 46.23 / 41.5 \\
        \bottomrule
    \end{tabular}
\end{table}

Precision does not rise noticeably on unanimous items (Table~\ref{tab:unanimous-precision}), suggesting that ambiguous items alone do not explain the low precision. This analysis does not establish a human performance level.

\section{Cover-All Baseline}
\label{sec:app-cover-all}

A cover-all response reproduces every valid path without exclusion. Its recall assumes unbounded answer length. Table~\ref{tab:cover-all} shows this baseline and scores for the ten evaluated models across Base and Oracle settings.

\begin{table}[htbp]
    \caption{Cover-all baseline and model score ranges across Base and Oracle settings.}
    \label{tab:cover-all}
    \centering
    \small
    \setlength{\tabcolsep}{4pt}
    \begin{tabular}{lcc}
        \toprule
        \textbf{Metric} & \shortstack{\textbf{Cover-all}\\(no exclusion)} & \shortstack{\textbf{Ten evaluated}\\\textbf{models}} \\
        \midrule
        Precision & 41.4 & 27--49 \\
        Recall & 100 & 42--79 \\
        PAS & 57.6 & 30--56 \\
        \bottomrule
    \end{tabular}
\end{table}

\end{document}